\documentclass{article}

\usepackage[accepted]{arxiv_version}

\usepackage[utf8]{inputenc} 
\usepackage[T1]{fontenc}    
\usepackage{url}            
\usepackage{booktabs}       
\usepackage{amsfonts}       
\usepackage{nicefrac}       
\usepackage{microtype}      
\usepackage{xcolor}         
\usepackage{wrapfig}
\usepackage[square,numbers]{natbib}
\usepackage{amssymb,mathtools}

\newcommand{\eqdef}{\mathrel{\mathop:}=}

\usepackage{multirow}
\usepackage{makecell}
\usepackage{tabu}
\usepackage{wrapfig}
\usepackage{colortbl}
\usepackage{array}

\usepackage{amsmath,amsfonts,bm}

\def\eqref#1{equation~\ref{#1}}

\def\1{\bm{1}}

\DeclareMathAlphabet{\mathsfit}{\encodingdefault}{\sfdefault}{m}{sl}
\SetMathAlphabet{\mathsfit}{bold}{\encodingdefault}{\sfdefault}{bx}{n}

\DeclareMathOperator*{\argmax}{arg\,max}

\DeclareMathOperator{\sign}{sign}

\usepackage{xcolor}
\definecolor{mydarkred}{rgb}{0.6,0,0}
\definecolor{mydarkgreen}{rgb}{0,0.6,0}
\definecolor{mydarkblue}{rgb}{0,0,0}
\usepackage{url}
\usepackage{microtype}
\usepackage{graphicx}
\usepackage{subfigure}
\usepackage{booktabs}
\usepackage{amsthm}
\usepackage{algorithm}
\usepackage{algorithmic}
\usepackage{soul} 
\definecolor{mydarkred}{rgb}{0.6,0,0}
\definecolor{mydarkgreen}{rgb}{0,0.6,0}
\definecolor{mydarkblue}{rgb}{0,0,0.6}
\usepackage[colorlinks,
linkcolor=mydarkred,
citecolor=mydarkgreen]{hyperref}

\usepackage{amsthm}
\newtheorem{mydef}{Definition}

\title{Local Reweighting for Adversarial Training}
\begin{document}

\onecolumn
{\centering \Large \textbf{Local Reweighting for Adversarial Training}}\\
\\
\\
\textbf{Ruize Gao}$^{1,\dagger}$~~\textbf{Feng Liu}$^{2,\dagger}$~~\textbf{Kaiwen Zhou}$^{1}$~~\textbf{Gang Niu}$^{3}$~~\textbf{Bo Han}$^{4}$ \textbf{James Cheng}$^{1}$\\
\\
{
\small
$^{1}$Department of Computer Science and Engineering, The Chinese University of Hong Kong, HKSAR, China\\
$^{2}$DeSI Lab, Australian AI Institute, University of Technology Sydney, Sydney, Australia\\
$^{3}$RIKEN Center for
Advanced Intelligence Project (AIP), Tokyo, Japan\\
$^{4}$Department of Computer Science, Hong Kong Baptist University, HKSAR, China\\
$^{\dagger}$Equal Contribution\\}
{\small ~~~~Emails: sjtu16.brian.gao@gmail.com, feng.liu@uts.edu.au, kwzhou@cse.cuhk.edu.hk, gang.niu@riken.jp, 
\\
{~~~~~~~~~~~~~bhanml@comp.hkbu.edu.hk, jcheng@cse.cuhk.edu.hk}}

\rule{\textwidth}{0.4mm}\\

{\large \textbf{Abstract}}
\vspace{1em}

\emph{Instances-reweighted adversarial training}~(IRAT) can significantly boost the robustness of trained models, where data being less/more vulnerable to the \emph{given} attack are assigned smaller/larger weights during training.
However, when tested on attacks \emph{different from} the given attack simulated in training, the robustness may drop significantly (e.g., even worse than no reweighting).
In this paper, we study this problem and propose our solution---\emph{locally reweighted adversarial training}~(LRAT).
The rationale behind IRAT is that we do not need to pay much attention to an instance that is already safe under the attack.
We argue that the safeness should be \emph{attack-dependent}, so that for the same instance, its weight can change given different attacks based on the same model.
Thus, if the attack simulated in training is \emph{mis-specified}, the weights of IRAT are misleading.
To this end, LRAT \emph{pairs} each instance with its adversarial variants and performs \emph{local reweighting inside each pair}, while performing \emph{no global reweighting}---the rationale is to fit the instance itself if it is immune to the attack, but not to skip the pair, in order to \emph{passively} defend different attacks in future.
Experiments show that LRAT works better than both IRAT (i.e., global reweighting) and the standard AT (i.e., no reweighting) when trained with an attack and tested on different attacks.

\rule{\textwidth}{0.4mm}\\
\section{Introduction}
A growing body of research shows that neural networks are vulnerable to adversarial examples, i.e., test inputs that are modified slightly yet strategically to cause misclassification \citep{carlini2017adversarial,finlayson2019adversarial,gao2020maximum,kurakin2016adversarial,nguyen2015deep,szegedy2013intriguing,wang2019convergence,zhang2020dual}. It is crucial to train a robust neural network to defend against such examples for security-critical computer vision systems, such as autonomous driving and medical diagnostics \citep{chen2015deepdriving,ma2021understanding,miyato2016adversarial,nguyen2015deep,szegedy2013intriguing}. To mitigate this issue, adversarial training methods have been proposed in recent years \citep{balunovic2019adversarial,goodfellow2014explaining, Madry18PGD,shafahi2020universal,wang2020once,wu2020adversarial}. By injecting adversarial examples into training data, adversarial training methods seek to train an adversarial-robust deep network whose predictions are locally invariant in a small neighborhood of its inputs \citep{bai2019hilbert,he2018decision,papernot2016towards,raghunathan2020understanding,tsipras2018robustness,yang2020closer}. 


Due to the diversity and complexity of adversarial data, over-parameterized deep networks have insufficient model capacity in \emph{adversarial training} (AT) \cite{zhang2020geometry}. To obtain a robust model given fixed model capacity, \citet{zhang2020geometry} suggest that we do not need to pay much attention to an instance that is already safe under the attack, and propose \emph{instance-reweighted adversarial training} (IRAT), which performs \emph{global reweighting} with a \emph{given} attack. To identify safe/non-robustness instances, they propose a \emph{geometric distance} between natural data points and current class boundary. Instances being closer to/farther from the class boundary is more/less vulnerable to the given attack, and should be assigned larger/smaller weights during AT. This \emph{geometry-aware IRAT} (GAIRAT) significantly boosts the robustness of the trained models when facing the given attack.
\begin{figure*}[tp]
    \begin{center}
        \subfigure[The weight distribution]
        {\includegraphics[width=0.329\textwidth]{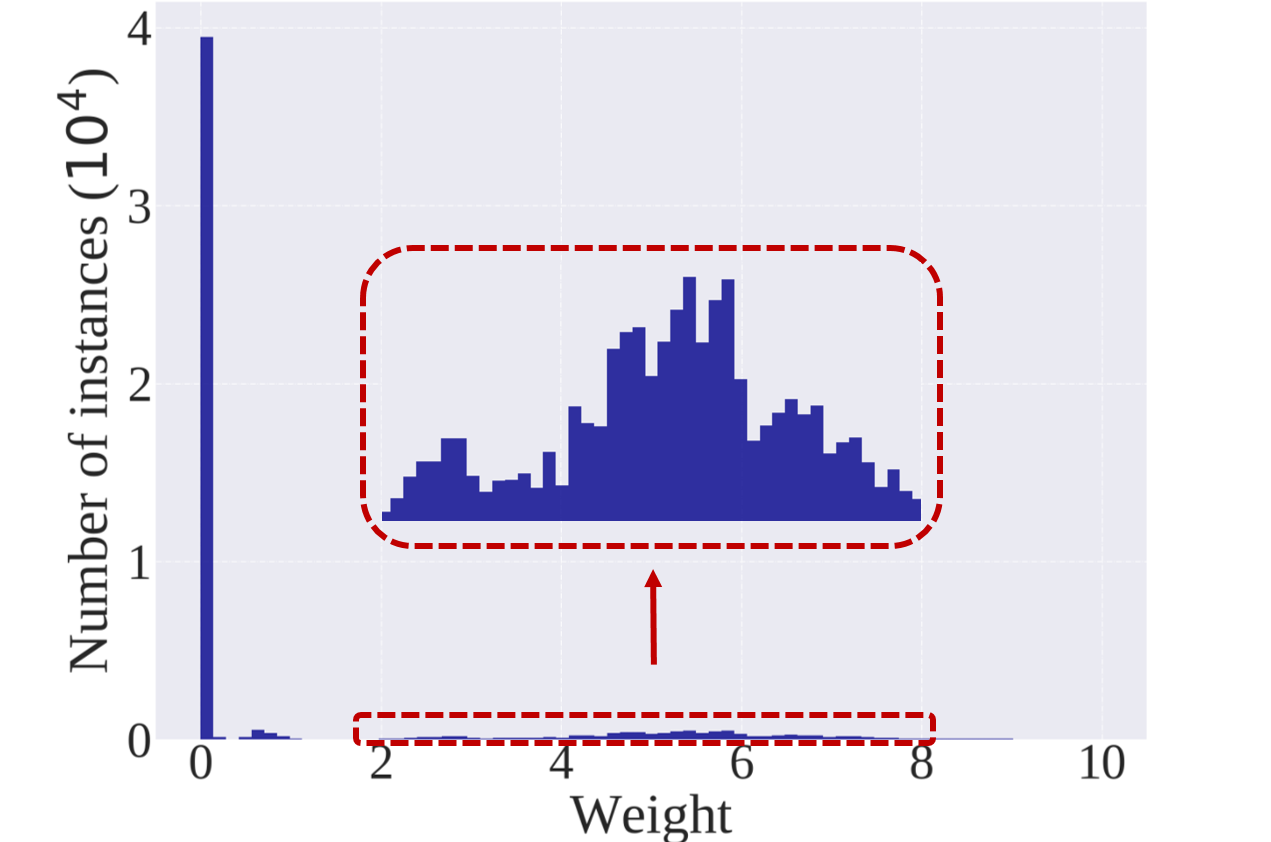}}
        \subfigure[Performance of SAT]
        {\includegraphics[width=0.329\textwidth]{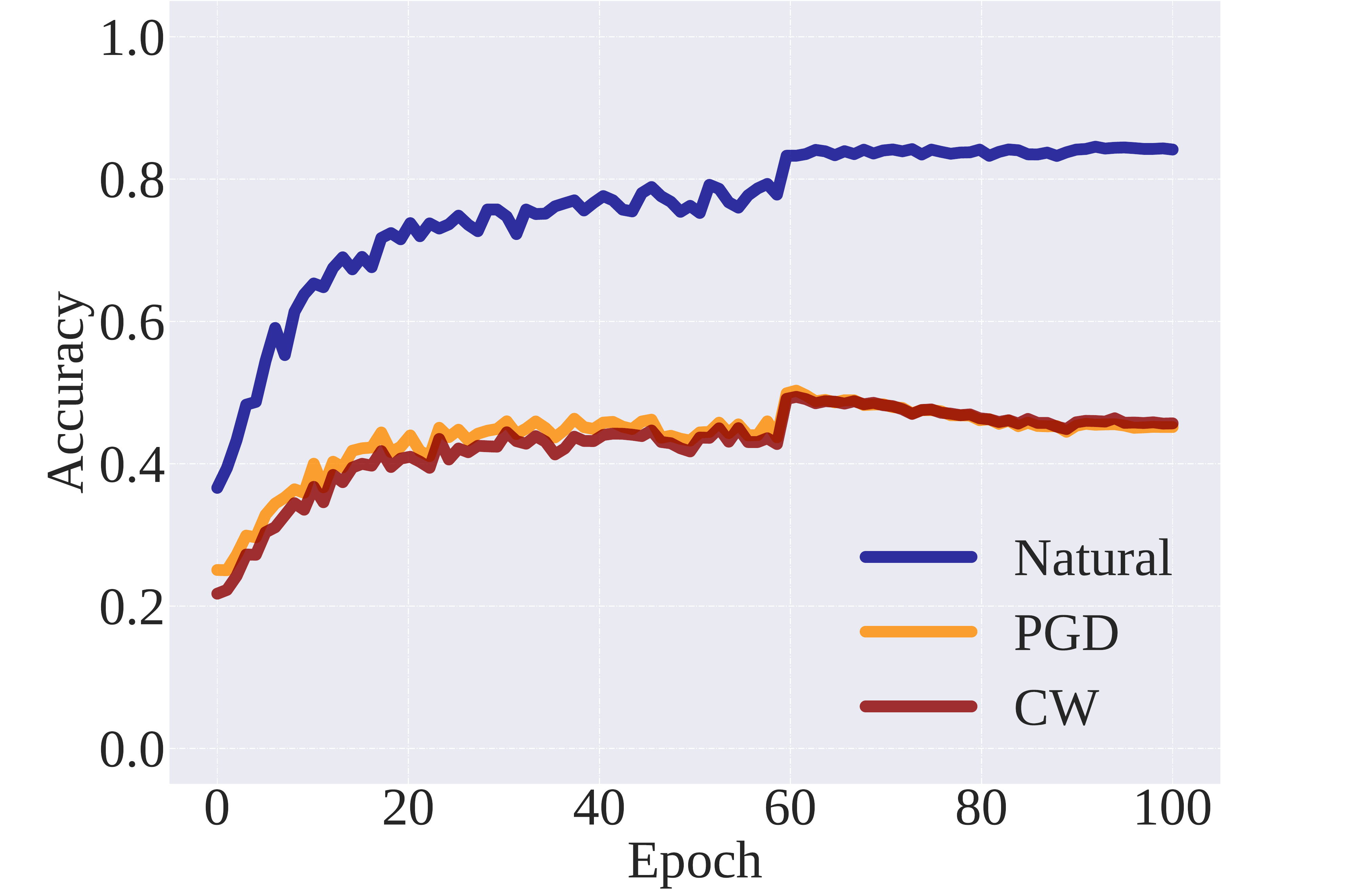}}
        \subfigure[Performance of GAIRAT]
        {\includegraphics[width=0.329\textwidth]{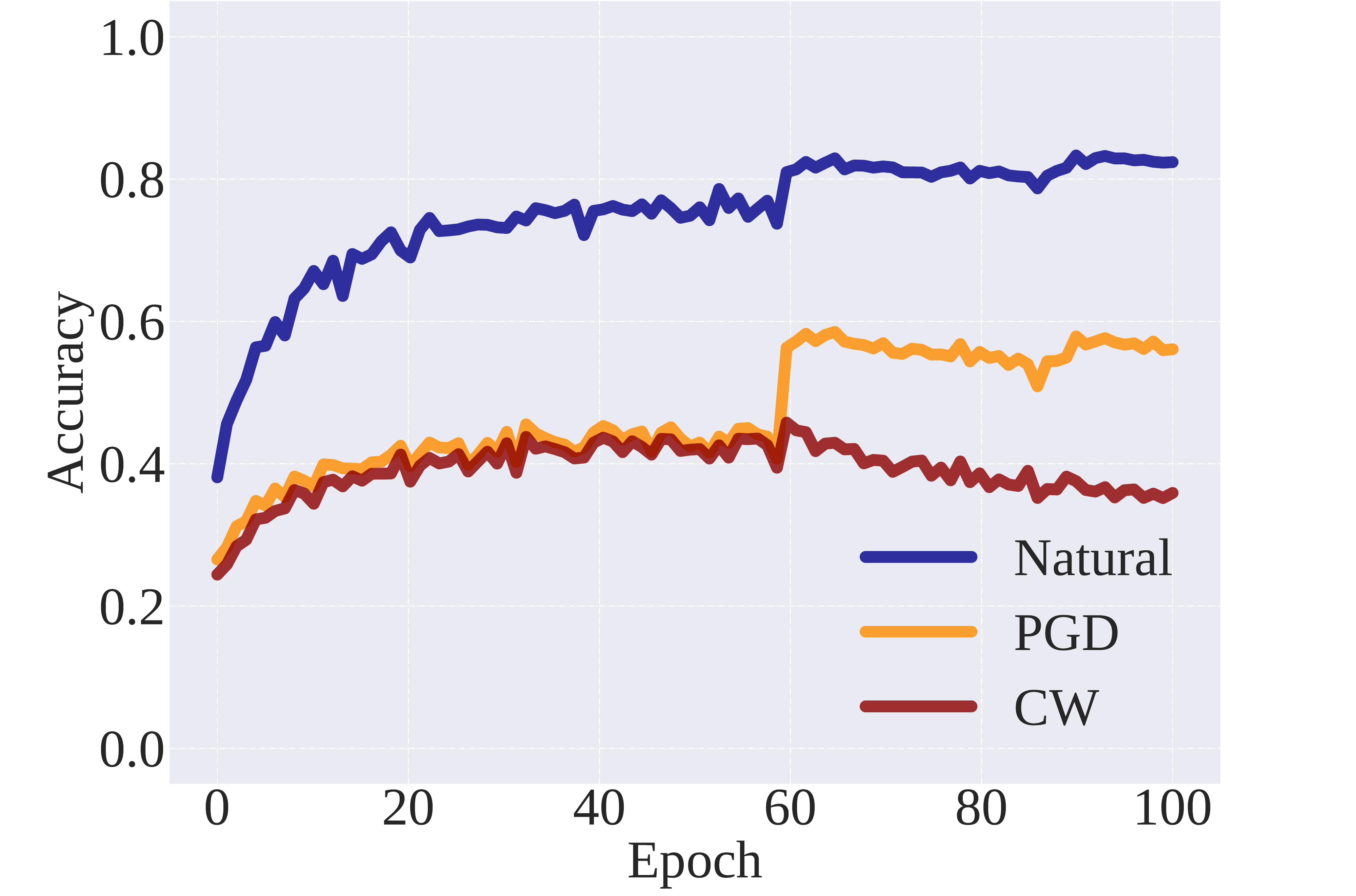}}
        \caption{\footnotesize
        The extreme reweighting in GAIRAT results in a significant decrease when facing an unseen attack. The subfigure (a) illustrates the frequency distribution of weights in the GAIRAT model on the CIFAR-10 training set, where approximately four-fifths of the instances are assigned very low weights (less than 0.2). The subfigure (b) and (c) illustrate the performance on the GAIRAT trained model and the AT trained model under different attacks, which show that GAIRAT does improve the robustness when attacked by PGD (used during training), but reduces the robustness when attacked by CW (unseen during training).
        }
    \label{fig:moti_1}
    \end{center}
    \vspace{-1em}
\end{figure*}

However, when tested on attacks that are \emph{different from} the given attack simulated in IRAT, the robustness of IRAT drops significantly (e.g., even worse than no reweighting). First, we find that a large number of instances are actually \emph{overlooked} during IRAT. 
Figure~\ref{fig:moti_1}(a) shows that, for approximately four-fifths of the instances, their corresponding adversarial variants are assigned very low weights (less than $0.2$). Second, we find that the robustness of the IRAT-trained classifier \emph{drops significantly} when facing an \emph{unseen} attack. Figures~\ref{fig:moti_1}(b) and \ref{fig:moti_1}(c) show that the classifier trained by GAIRAT (with \emph{projected gradient descent} (PGD) attack \citep{Madry18PGD}) has lower robustness when attacked by the unseen Carlini and Wagner attack (CW) \citep{carlini2017towards} compared with the robustness of \emph{standard adversarial training} (SAT) \citep{Madry18PGD}. 

\begin{figure*}[tp]
    \begin{center}
        \subfigure[$\mathcal{V}$-Union]
        {\includegraphics[width=0.23\textwidth]{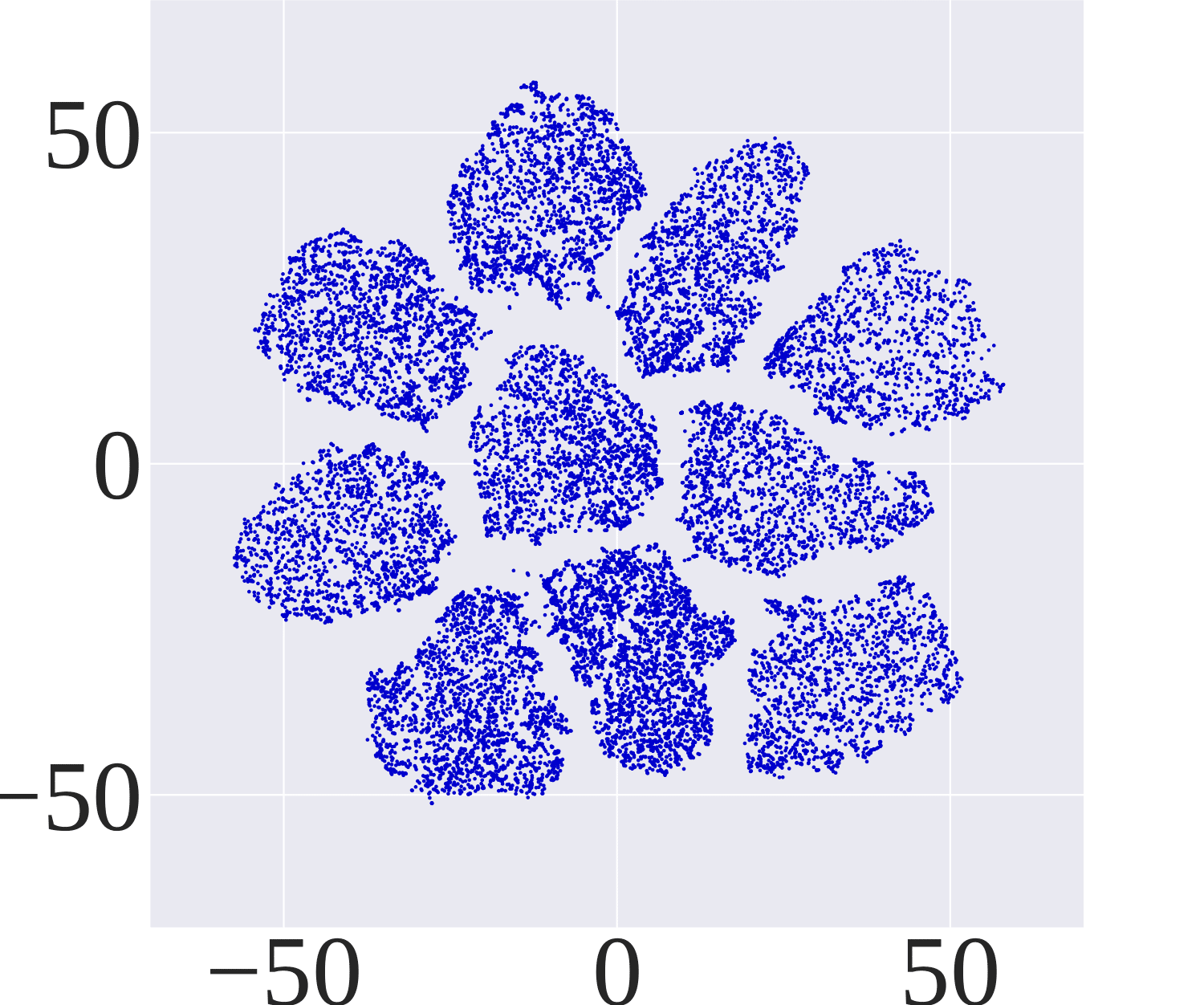}}
        \subfigure[$\mathcal{V}^{\textnormal{GD}}$-$\mathcal{V}^{\textnormal{CW}}$]
        {\includegraphics[width=0.23\textwidth]{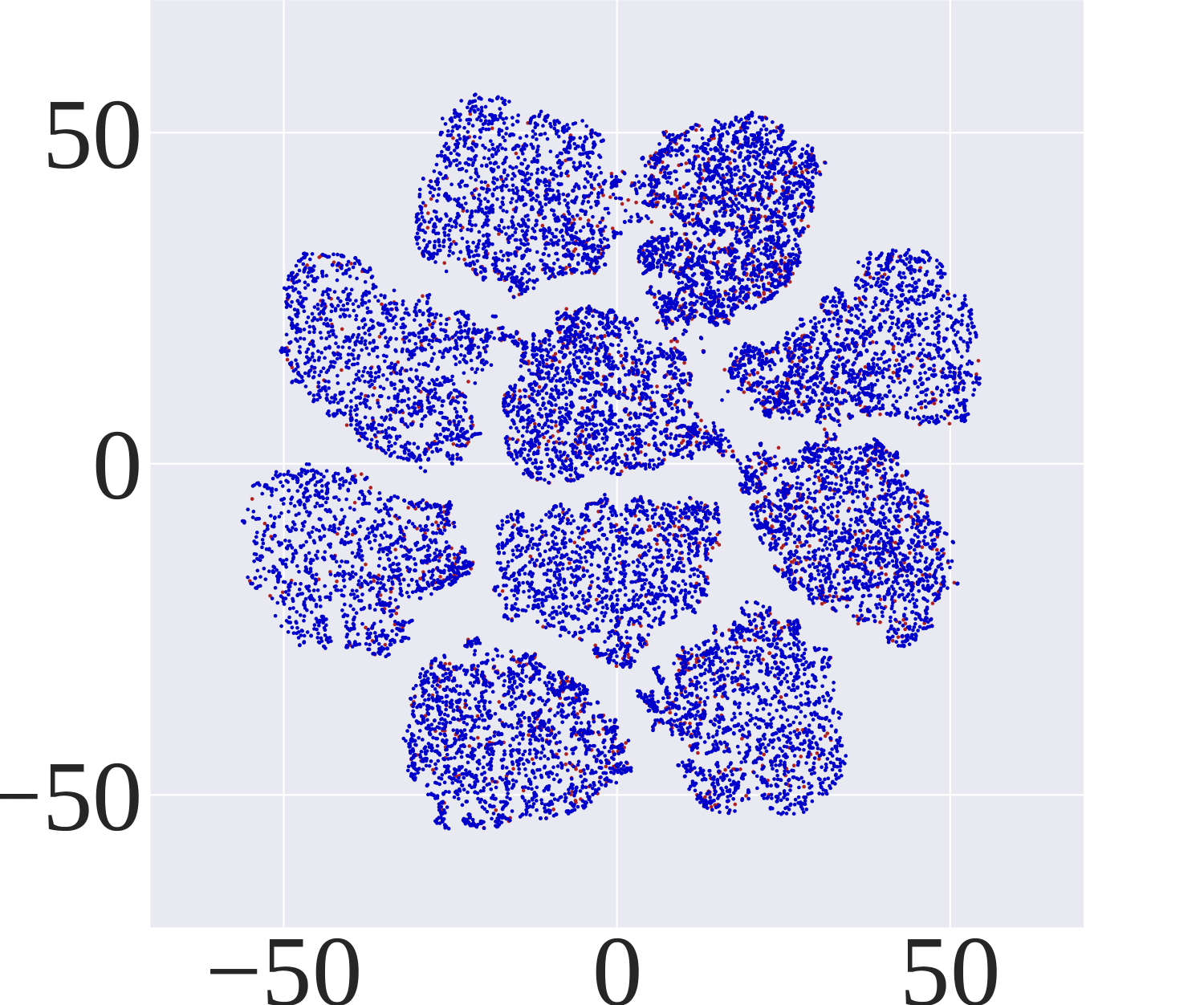}}
        \subfigure[$\mathcal{V}^{\textnormal{GD}}$-$\mathcal{V}^{\textnormal{PGD}}$]
        {\includegraphics[width=0.23\textwidth]{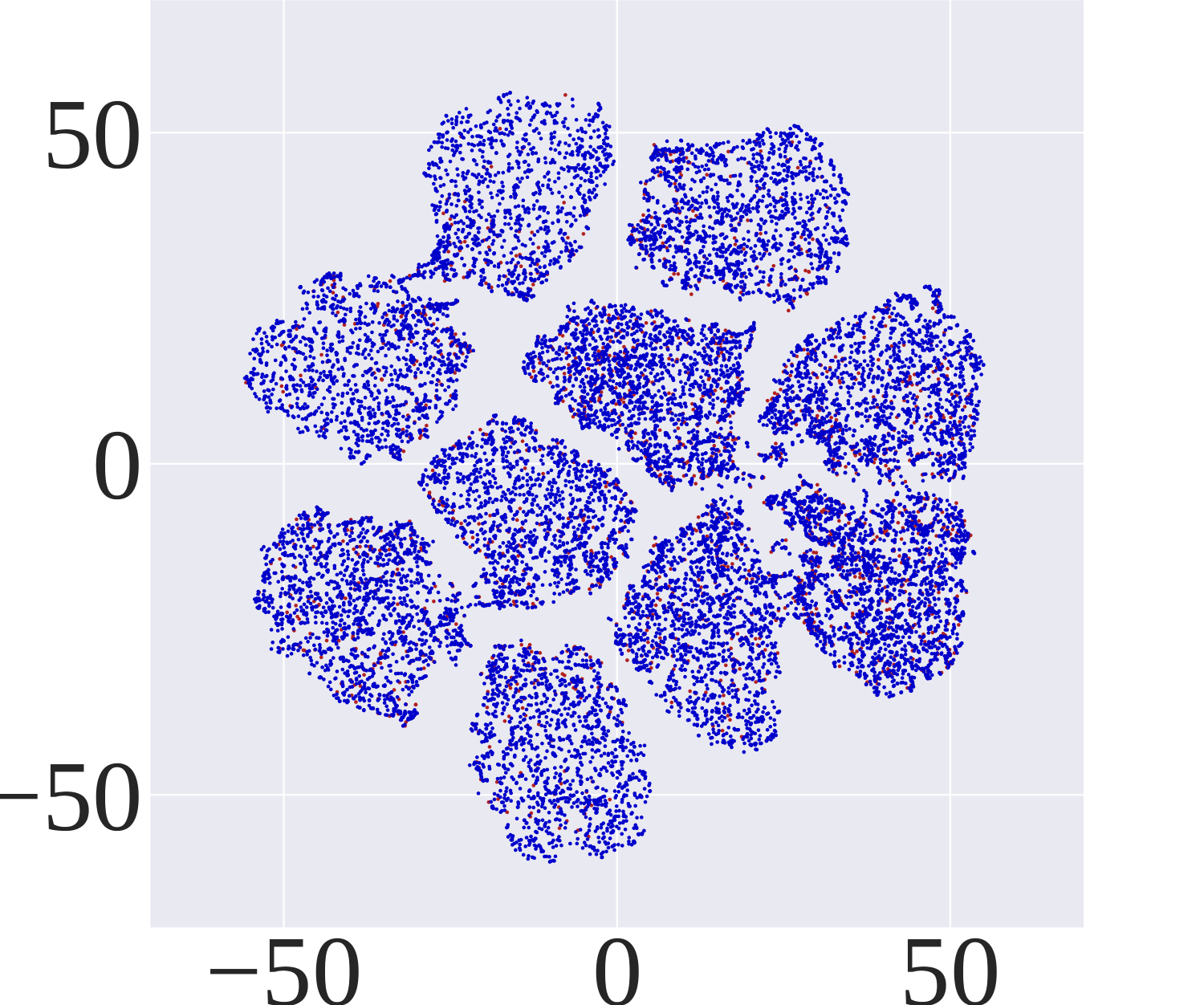}}
        \subfigure[$\mathcal{V}^{\textnormal{PGD}}$-$\mathcal{V}^{\textnormal{CW}}$]
        {\includegraphics[width=0.23\textwidth]{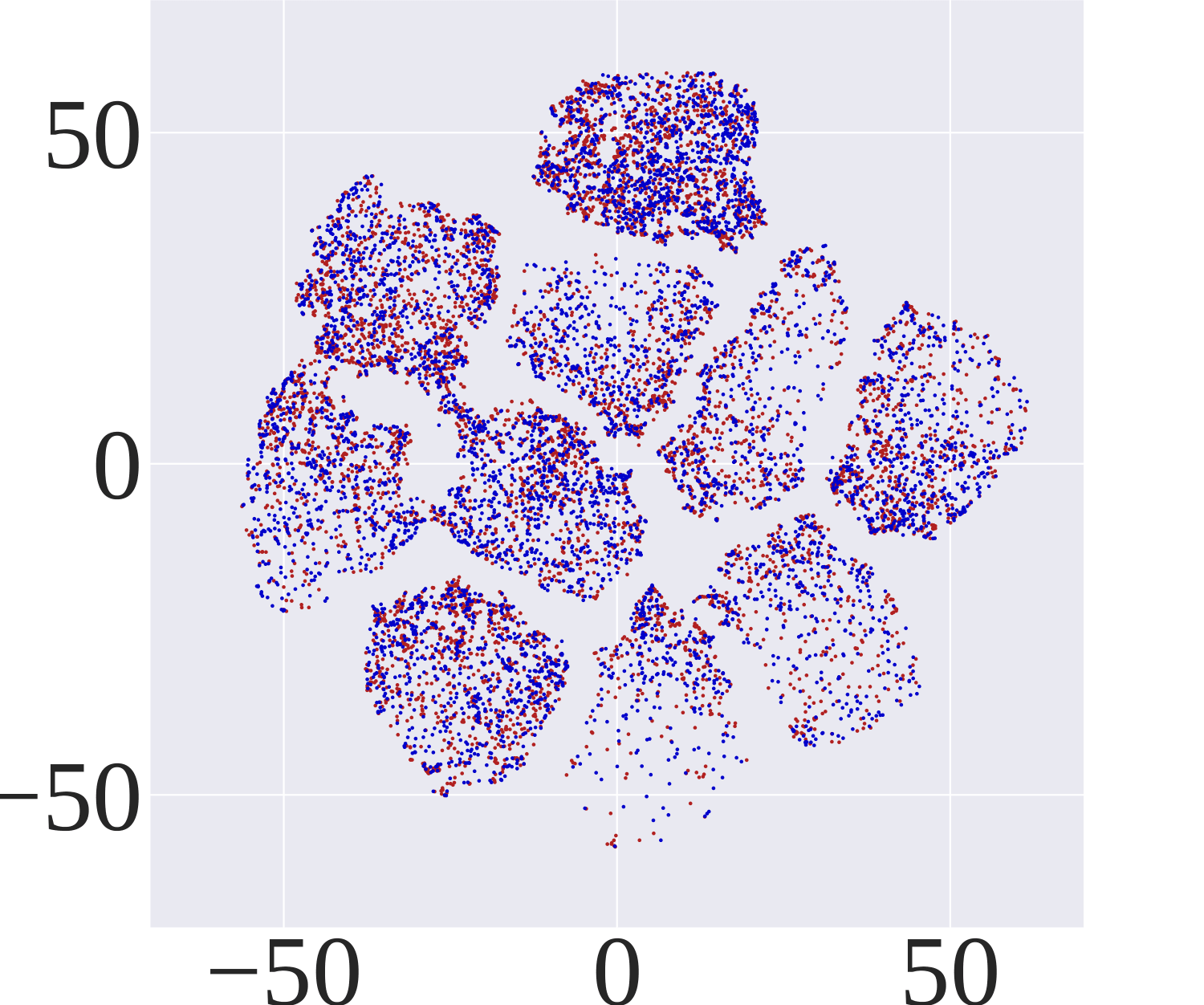}}
        \subfigure[$\mathcal{V}$-Union]
        {\includegraphics[width=0.23\textwidth]{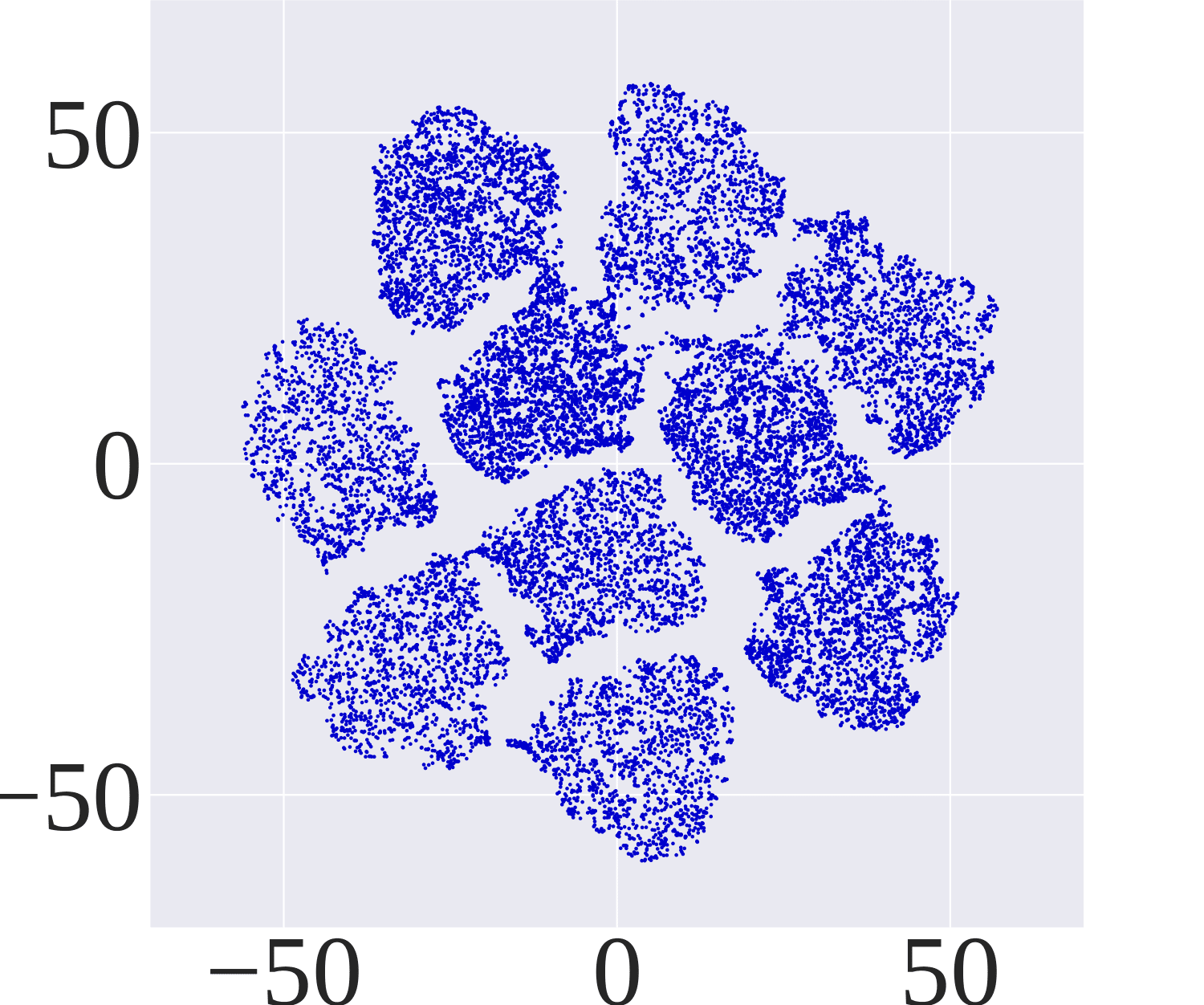}}
        \subfigure[$\mathcal{V}^{\textnormal{GD}}$-$\mathcal{V}^{\textnormal{CW}}$]
        {\includegraphics[width=0.23\textwidth]{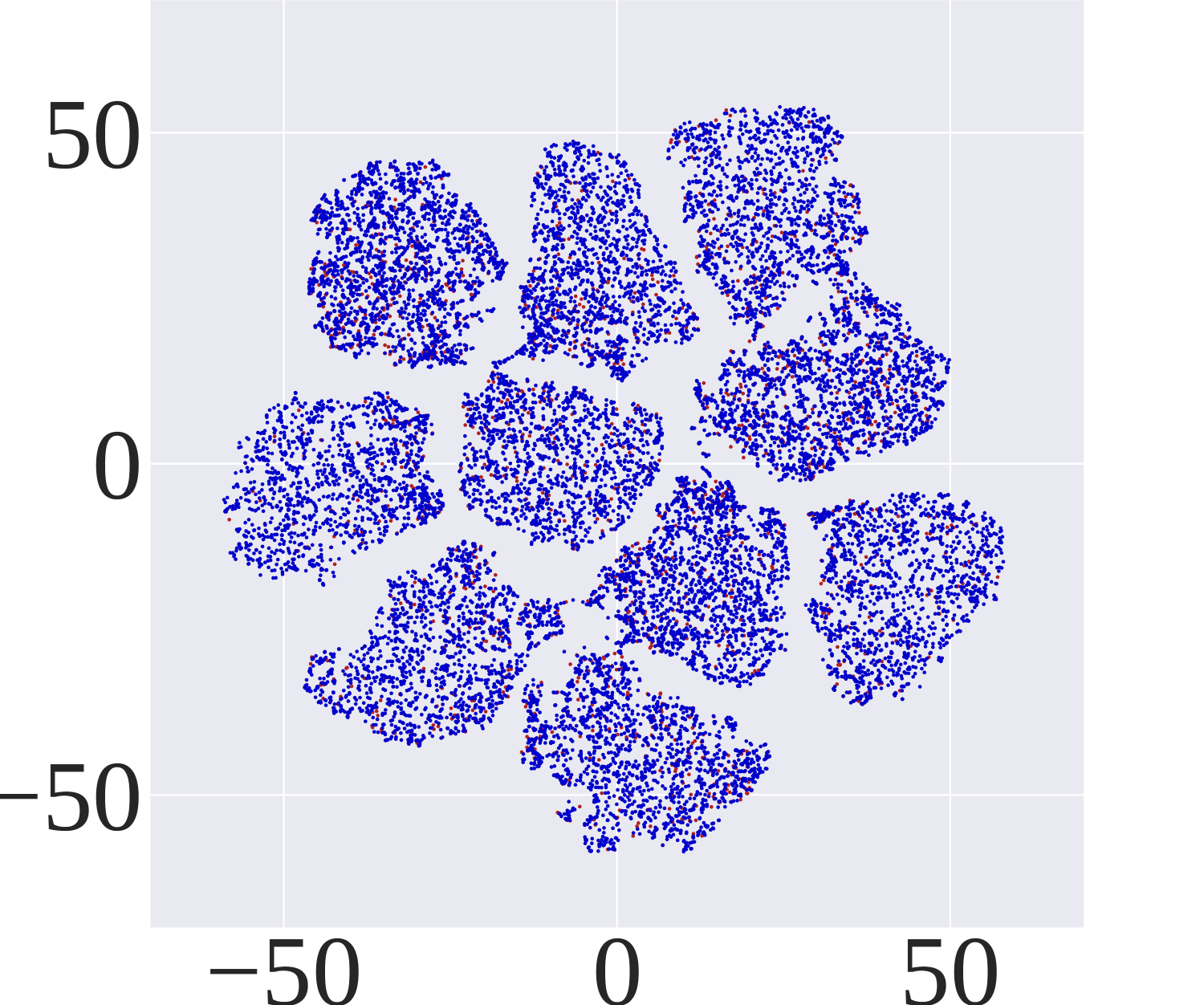}}
        \subfigure[$\mathcal{V}^{\textnormal{GD}}$-$\mathcal{V}^{\textnormal{PGD}}$]
        {\includegraphics[width=0.23\textwidth]{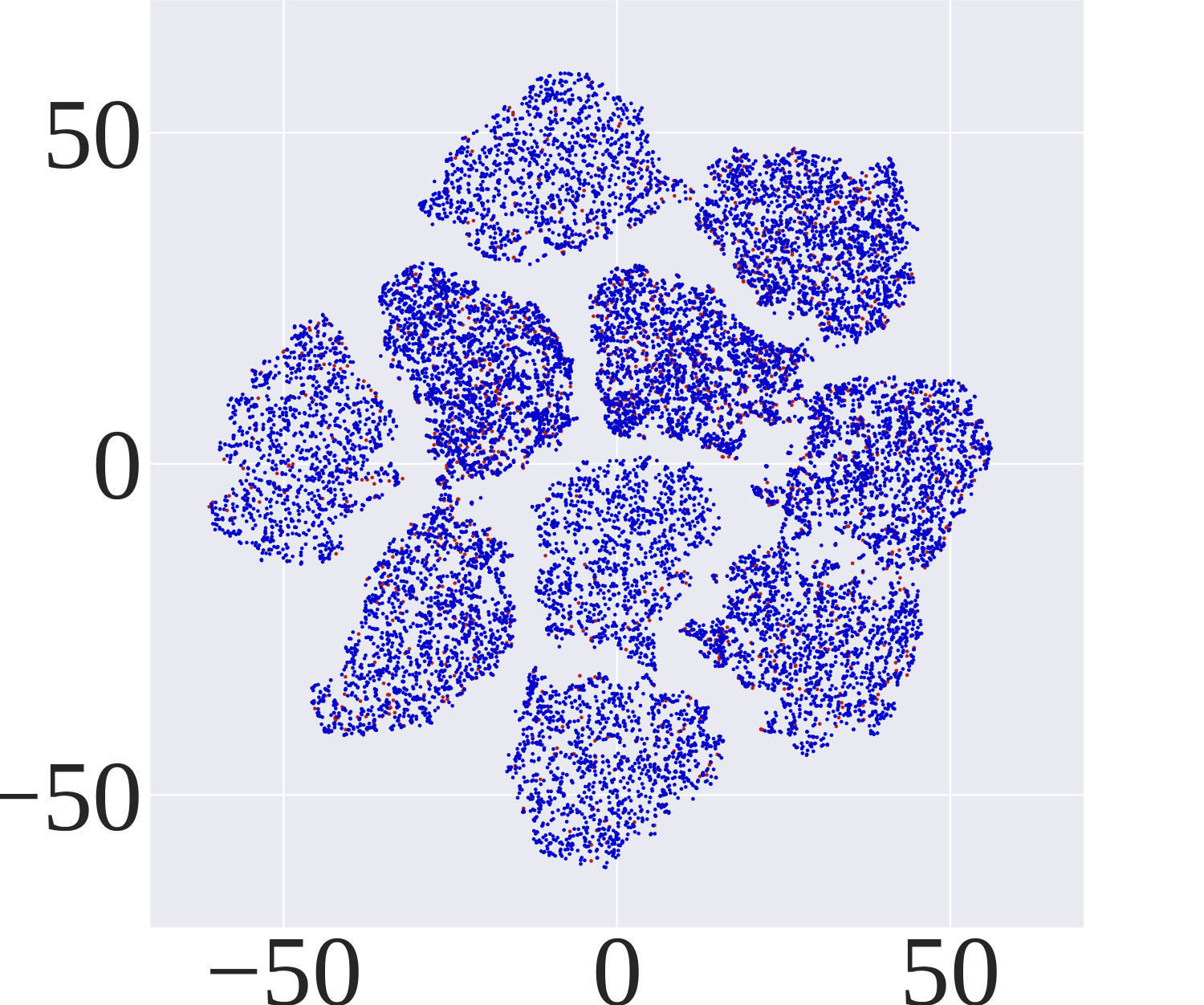}}
        \subfigure[$\mathcal{V}^{\textnormal{PGD}}$-$\mathcal{V}^{\textnormal{CW}}$]
        {\includegraphics[width=0.23\textwidth]{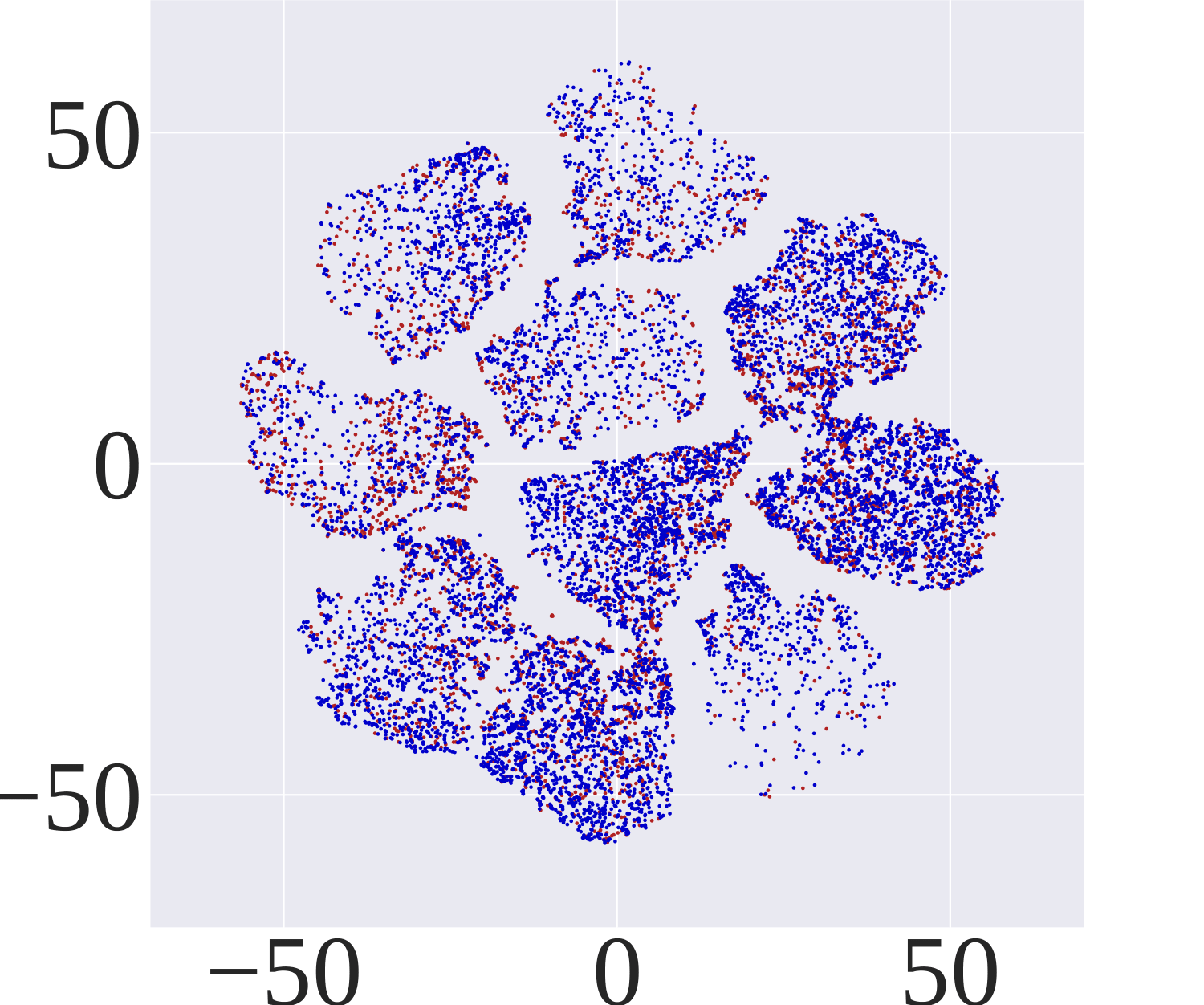}}
        \caption{\footnotesize
        Visualization of inconsistent vulnerable instances in different views using t-SNE. The dots in subfigures are outputs of the last layers in SAT-trained classifier (subfigures (a)-(d)) and GAIRAT-trained classifier (subfigures (e)-(h)). Classifier inputs are natural data points in the CIFAR-10 training set. The dots represent the vulnerable instances (top-20\%). The red dots represent consistently vulnerable instances, while the blue dots represent the inconsistent vulnerable instances in the view of $\mathcal{V}^{\textnormal{GD}}$, $\mathcal{V}^{\textnormal{PGD}}$ and $\mathcal{V}^{\textnormal{CW}}$. $\mathcal{V}^{\textnormal{GD}}$ is the vulnerability w.r.t. \emph{geometry distance} (GD) defined by \citet{zhang2020geometry}. $\mathcal{V}^{\textnormal{PGD}}$ and $\mathcal{V}^{\textnormal{CW}}$ are the vulnerability measurement function w.r.t. PGD and CW defined in Eq.~(\ref{omega:PGD}) and Eq.~(\ref{omega:CW}), respectively. We can clearly see that there exist abundant blue dots but only a few red dots, which means a large number of inconsistent vulnerable instances in different views. 
        }
    \label{fig:moti_tsne}
    \end{center}
    \vspace{-0.7em}
\end{figure*}

In this paper, we investigate the reasons behind this phenomenon. Unlike the common scenario of the classification problem where the training and testing data are fixed, there are different adversarial variants for the same instance in AT, e.g., PGD-based or CW-based adversarial variants. A natural question comes with this---whether there are inconsistent vulnerabilities in the view of different attacks? The answer is \emph{affirmative}. Figure~\ref{fig:moti_tsne} visualizes this phenomenon using t-SNE \citep{van2008visualizing}. The $8$ subfigures in Figure~\ref{fig:moti_tsne} visualize inconsistent vulnerable instances in different views using t-SNE. The red dots in all subfigures represent consistently vulnerable instances between different views, while the blue dots represent the inconsistent vulnerable instances. From both the SAT-trained classifier (Figure~\ref{fig:moti_tsne}(a)-\ref{fig:moti_tsne}(d)) and GAIRAT-trained classifier (Figure~\ref{fig:moti_tsne}(e)-\ref{fig:moti_tsne}(h)), we can clearly see that a large number of vulnerable instances are inconsistent (the blue dots dominate in all the subfigures).


Given the above investigation, we argue that the safeness of instances is \emph{attack-dependent}, that is, for the same instance, its weight can change given different attacks based on the same model. Thus, if the attack simulated in training is \emph{mis-specified}, the weights of IRAT are misleading. In order to ameliorate this pessimism of IRAT, we propose our solution---\emph{locally reweighted adversarial training}~(LRAT). As shown in Figure~\ref{fig:sketch}, LRAT \emph{pairs} each instance with its adversarial variants and performs \emph{local reweighting inside each pair}, while performing \emph{no global reweighting}. The rationale of LRAT is to fit the instance itself if it is immune to the attack, and in order to \emph{passively} defend different attacks in future, LRAT does not skip the pair. For the realization of LRAT, we propose a general \emph{vulnerability-based} reweighting strategy that is applicable to various attacks instead of the geometric distance that is only compatible with the PGD attack \cite{zhang2020geometry}.


Our experimental results show that LRAT works better than both SAT (i.e., no reweighting) and IRAT (i.e., global reweighting) when trained with an attack but tested on different attacks. For other existing adversarial training methods, e.g., TRADES \citep{ZhangYJXGJ19TRADES}, we also design LRAT for TRADES \citep{ZhangYJXGJ19TRADES} (i.e., LRAT-TRADES). Our results also show that LRAT-TRADES works better than both TRADES and IRAT-TRADES.
\newcommand{\T}{{\hspace{-0.25ex}\top\hspace{-0.25ex}}}
\newcommand{\ST}{\mathrm{s.t.}}
\newcommand{\bE}{\mathbb{E}}
\newcommand{\bR}{\mathbb{R}}
\newcommand{\cB}{\mathcal{B}}
\newcommand{\cF}{\mathcal{F}}
\newcommand{\cX}{\mathcal{X}}
\newcommand{\cY}{\mathcal{Y}}
\newcommand{\bx}{{x}}
\newcommand{\bxprime}{{x}'}
\newcommand{\bxtidle}{\tilde{{x}}}
\newcommand{\epsball}{\mathcal{B}_\epsilon}
\newcommand{\yadv}{\tilde{y}}
\section{Adversarial Training}
\label{Sec:Preli}
In this section, we briefly review existing adversarial training methods \citep{Madry18PGD, zhang2020geometry}.
Let $(\cX,d_\infty)$ be the input feature space $\cX$ with a metric $d_{\infty}(x,x^{\prime})=\|x-x^{\prime}\|_\infty$, and $\epsball[x] = \{x^{\prime}  \mid d_{\infty}(x,x')\le\epsilon\}$
be the closed ball of radius $\epsilon>0$ centered at $x$ in $\cX$. The dataset $S = \{ x_i, y_i\}^n_{i=1}$, where $x_i \in \cX$, $y_i\in\mathcal{Y} =\{0,1,\dots,K-1\}$. We use $f_{\theta}(x)$ to denote a deep neural network parameterized by $\theta$. Specifically,
$f_\theta(x)$ predicts the label of an input instance $x$ via:
\begin{align}
f_\theta(x) =\mathop{\argmax}\limits_{k\in K}\mathrm{p}_k(x;\theta),
\label{Sec2:pt}
\end{align}
where $\mathrm{p}_k(x;\theta)$ denotes the predicted probability (softmax on logits) of $x$ belonging to class $k$.
\subsection{Standard Adversarial Training}
The objective function of SAT \citep{Madry18PGD} is
\begin{align}
\mathop{\min}\limits_{f_\theta \in \mathcal{F}}\frac{1}{n}\sum_{i=1}^n \ell(f_\theta(\tilde{x}_i),y_i),
\label{Equ-AT-1}
\end{align}
where $\tilde{x}_i =\mathop{\argmax}_{\tilde{x}_i \in \epsball[x]}\ell(f_\theta(\tilde{x}_i),y_i)$.
The selected $\tilde{x}$ is the most adversarial variant within the $\epsilon$-ball center at $x$. The loss function $\ell:\mathbb{R}^{K}\times \mathcal{Y} \to \mathbb{R}$ is a composition of a base loss $\ell_{B}:\bigtriangleup^{K-1}\times \mathcal{Y} \to \mathbb{R}$ (e.g., the cross-entropy loss) and an inverse link function $\ell_L :  \mathbb{R}^{K} \to \bigtriangleup^{K-1}$ (e.g., the soft-max activation), where $\bigtriangleup^{K-1}$ is the corresponding probalility simplex. Namely, $\ell(f_\theta(\cdot),y)=\ell_{B}(\ell_{L}(f_\theta(\cdot)),y)$.
PGD \citep{Madry18PGD} is the most common approximation method for searching the most adversarial variant. Starting from $x^{(0)} \in \cX$, PGD (with step size $\alpha > 0$) works as follows:
\begin{align}
x^{(t+1)} = \Pi_{\mathcal{B}_\epsilon[x^{(0)}]}(x^{(t)}+\alpha \sign(\nabla_{x^{t}}\ell(f_{\theta}(x^{(t)},y))), t \in \mathbb{N},
\label{Equ-AT-3}
\end{align}
where $\mathbb{N}$ is the number of iterations; $x^{(0)}$ refers to the starting point that natural instance (or a natural instance perturbed by a small Gaussian or uniformly random noise); $y$ is the corresponding label for $x^{(0)}$; $x^{(t)}$ is the adversarial variant at step $t$; $\Pi_{\mathcal{B}_\epsilon[x^{(0)}]}(\cdot)$ is the projection function that projects the adversarial variant back into the $\epsilon$-ball centered at $x^{(0)}$ if necessary. 

\subsection{Geometry-Aware Instance-Reweighted Adversarial Training}
GAIRAT is a typical IRAT proposed by \citet{zhang2020geometry}. GAIRAT argues that natural training data farther from/close to the decision boundary are safe/non-robustness, and should be assigned with smaller/larger weights. Let $\omega(x, y)$ be the geometry-aware weight assignment function on the loss of the adversarial variant $\tilde{x}$, where the generation of $\tilde{x}$ follows SAT. GAIRAT aims to
\begin{align}
\mathop{\min}\limits_{f_\theta \in \mathcal{F}}\frac{1}{n}\sum_{i=1}^n \omega(x_i, y_i)\ell(f_\theta(\tilde{x}_i),y_i).
\label{Equ-GAIRAT}
\end{align}
Eq.~(\ref{Equ-GAIRAT}) rescales the loss using a function $\omega(x, y)$. This function is non-increasing w.r.t. GD, which is defined as the least steps that the PGD method needs to successfully attack the natural instances.
The method then normalizes $\omega$ to ensure that $\omega(x, y) \geq 0$ and $\frac{1}{n}\sum_{i=1}^{n}\omega(x_i, y_i) = 1$.
Finally GAIRAT employs a bootstrap period in the initial part of the training by setting $\omega(x_i, y_i)=1$, thereby performing regular training and ignoring the geometric-distance of input $(x_i, y_i)$. 
\begin{figure}[tp]
    \begin{center}
        {\includegraphics[width=0.8\textwidth]{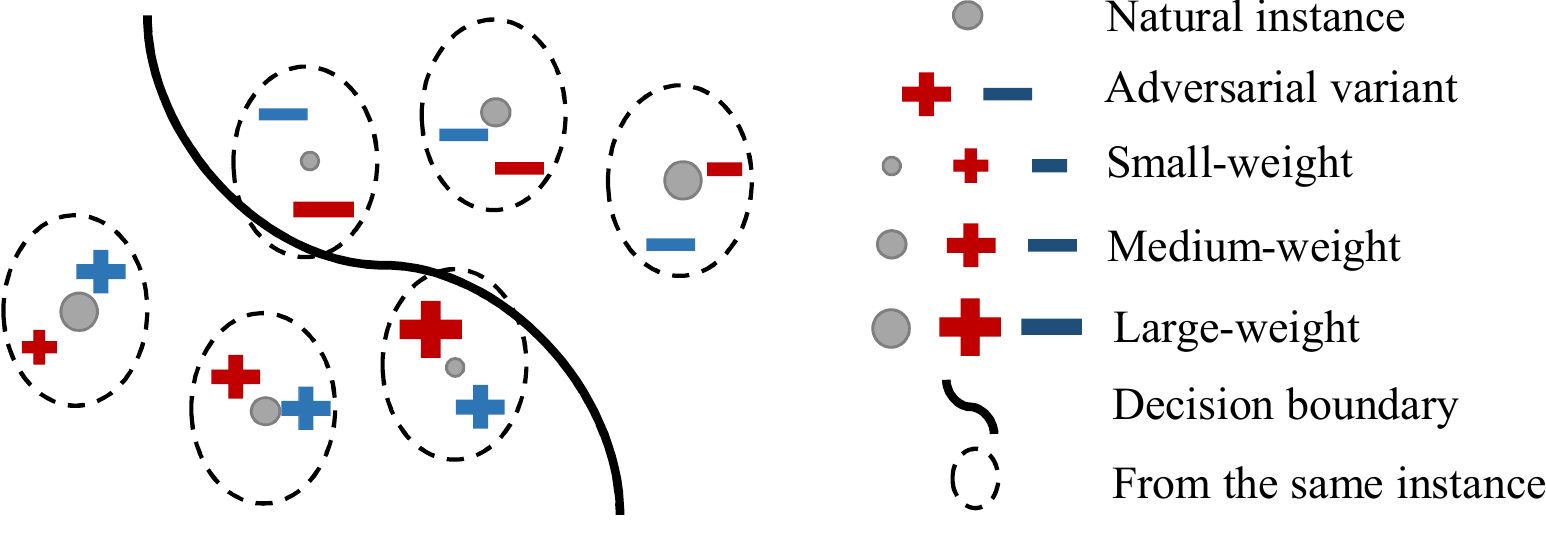}}
        \caption{\footnotesize\ 
        The illustration of LRAT. LRAT \emph{pairs} each instance with its adversarial variants, and performs \emph{local reweighting inside each pair} instead of global reweighting. For the same instance, there is inconsistent vulnerability in different views (the difference between red and blue). Thus, LRAT gives larger/smaller weights on the losses of adversarial variants, respectively.
        }
        \label{fig:sketch}
    \end{center}
    \vspace{-1.5em}
\end{figure}
\section{The Limitation of IRAT: Inconsistent Vulnerability in Different Views}
\label{GAIRAT_limitation}
\begin{figure*}[ht]
    \begin{center}
        \subfigure[All PGD]
        {\includegraphics[width=0.329\textwidth]{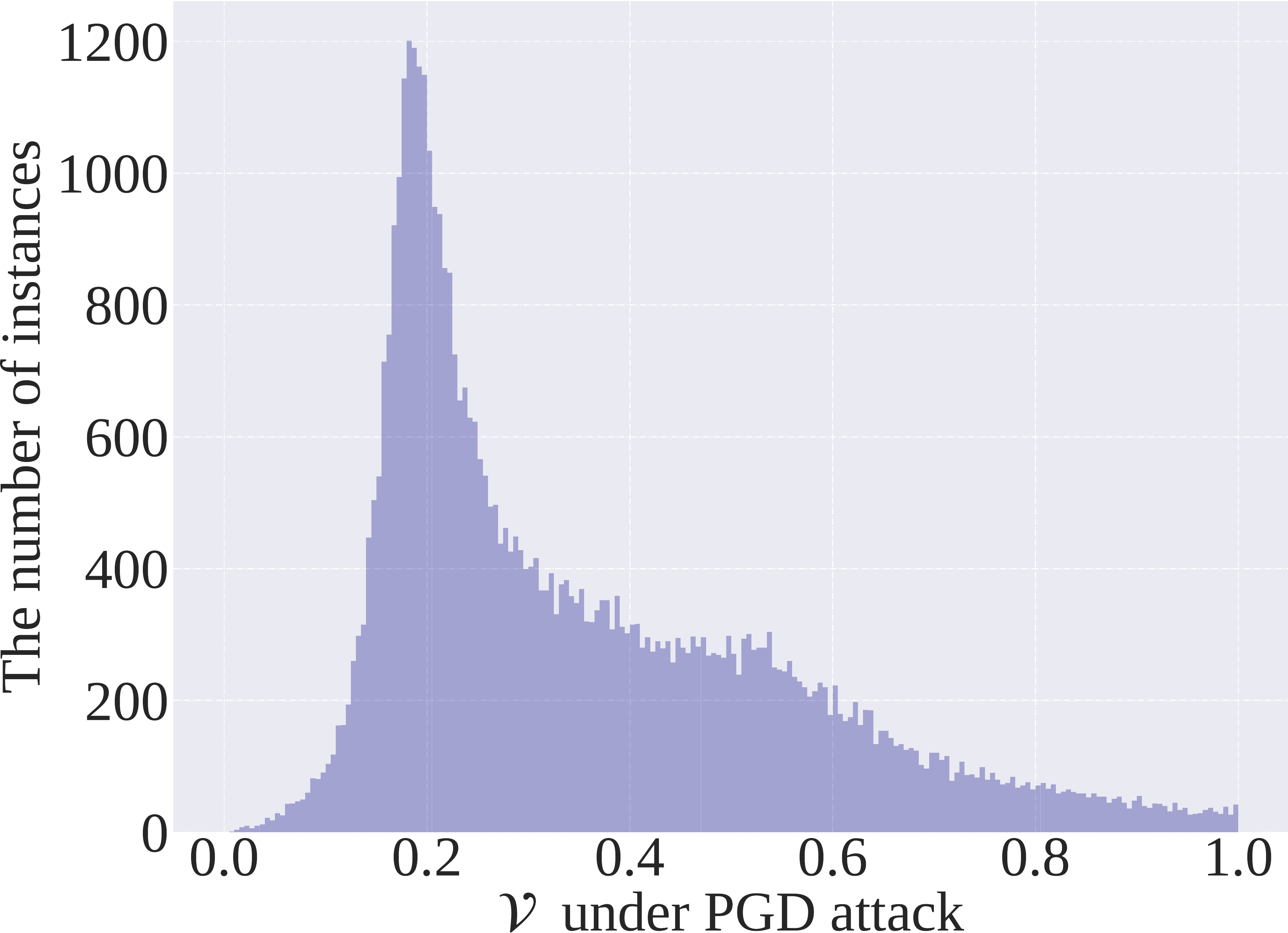}}
        \subfigure[Vulnerable PGD under all PGD]
        {\includegraphics[width=0.329\textwidth]{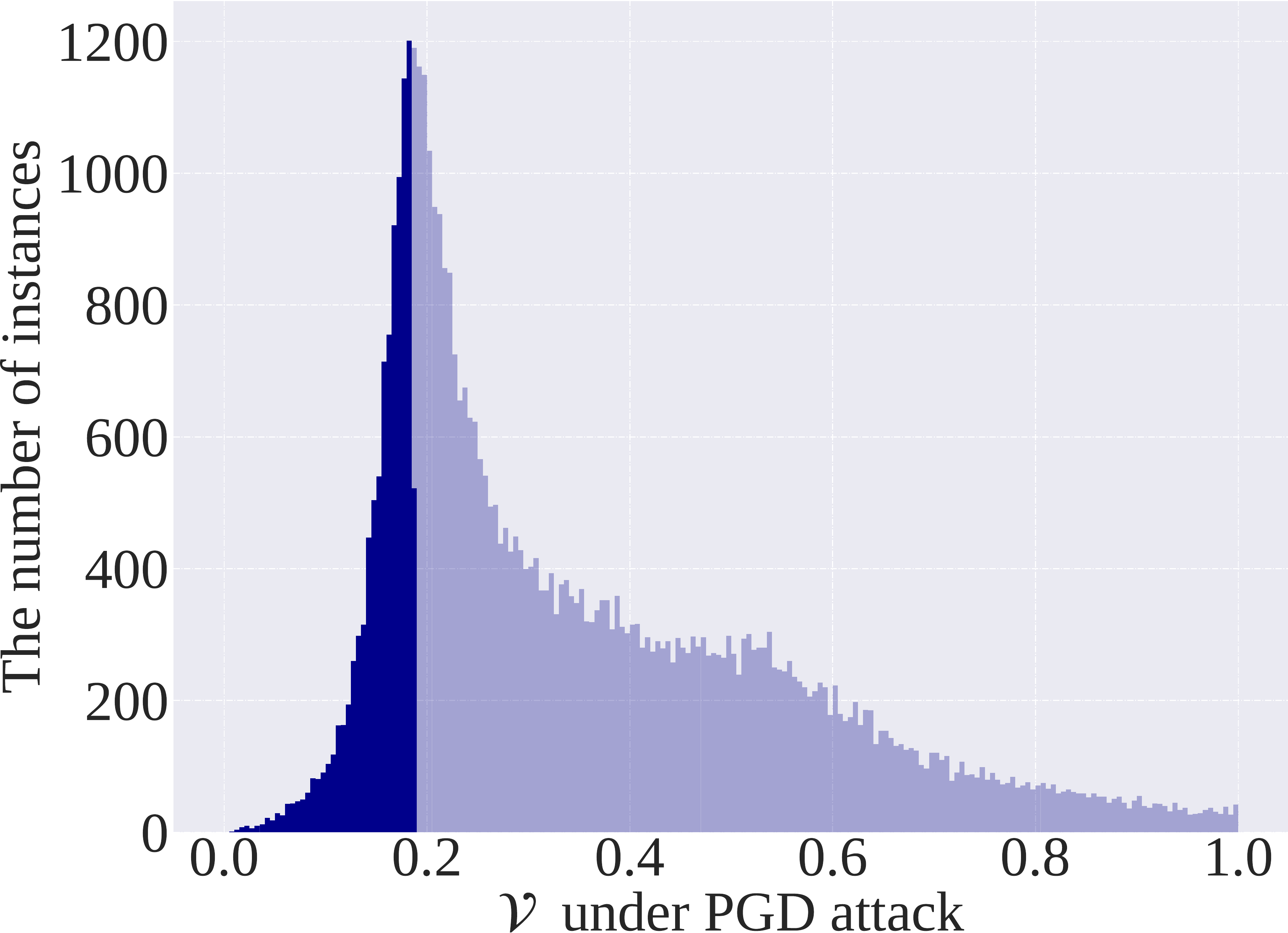}}
        \subfigure[Vulnerable CW under all PGD]
        {\includegraphics[width=0.329\textwidth]{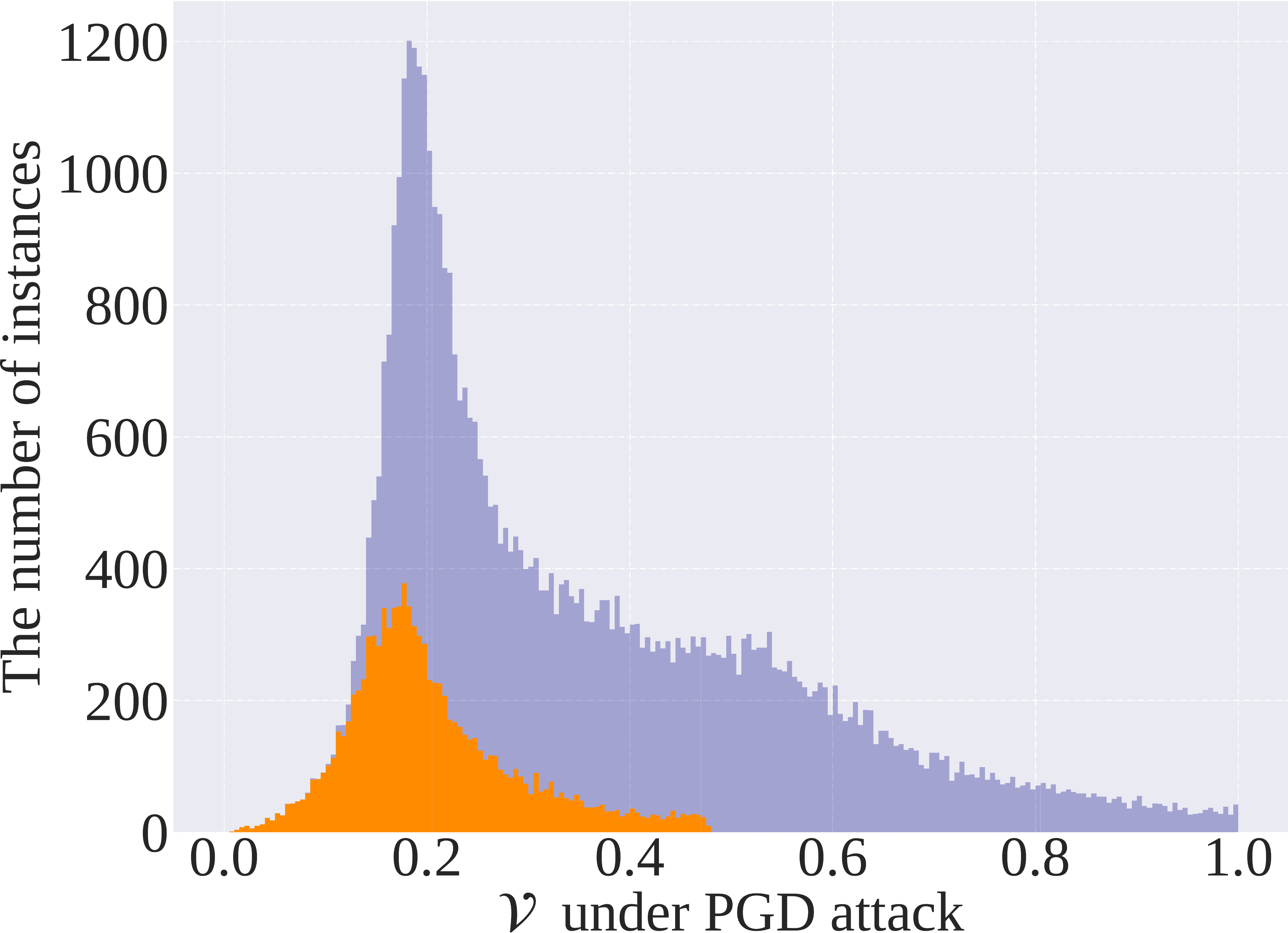}}
        \subfigure[All CW]
        {\includegraphics[width=0.329\textwidth]{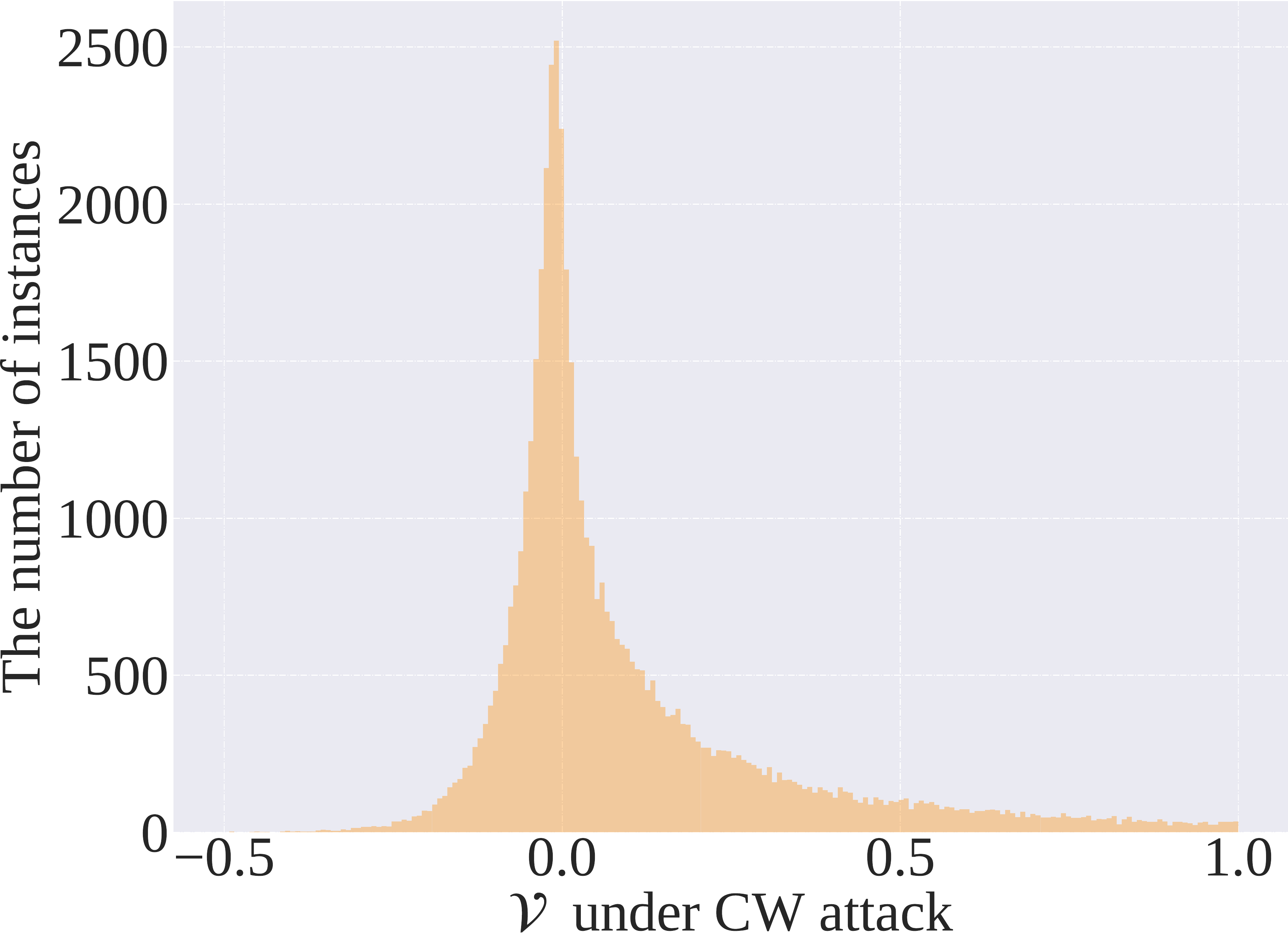}}
        \subfigure[Vulnerable CW under all CW]       
        {\includegraphics[width=0.329\textwidth]{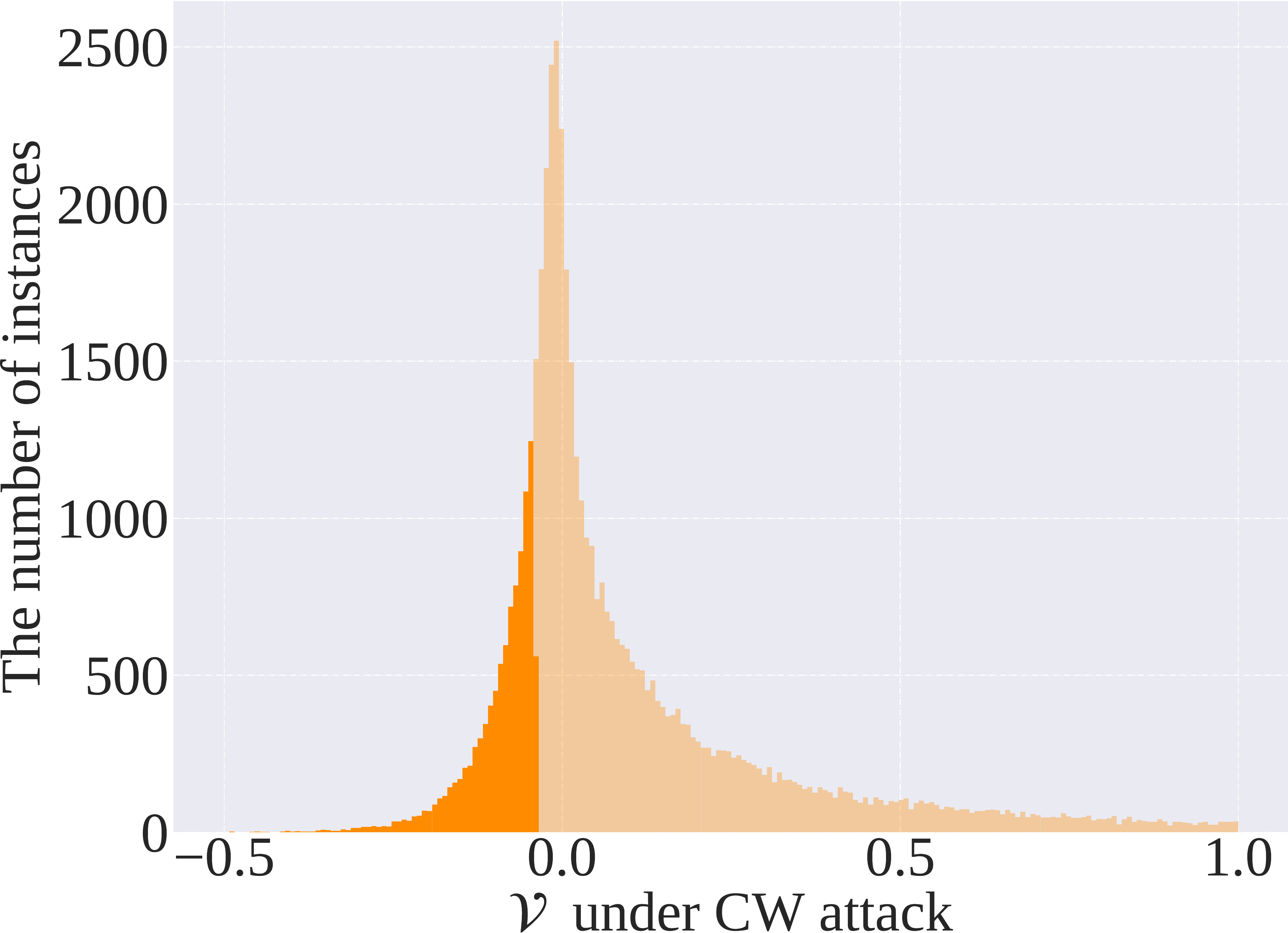}}
        \subfigure[Vulnerable PGD under all CW]
        {\includegraphics[width=0.329\textwidth]{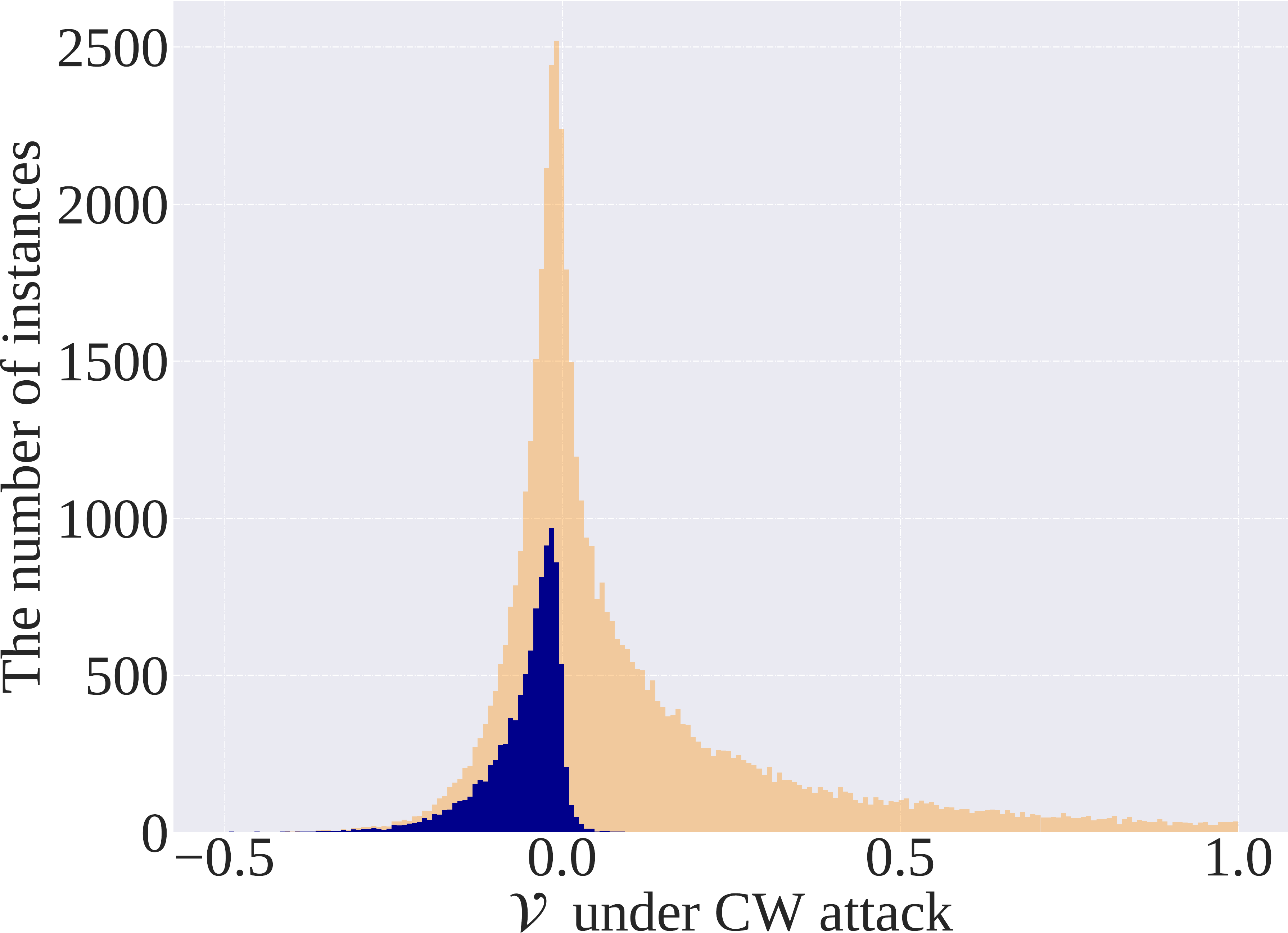}}
        \caption{\footnotesize
        The inconsistent vulnerability between instances about different variants. Six subfigures illustrate the frequency distribution of adversarial variants in the GAIRAT trained model on the CIFAR-10 training set (50,000). The x-coordinate in subfigures (a)-(c) is $\mathcal{V}$ in the view of PGD. The x-coordinate in subfigures (d)-(f) is $\mathcal{V}$ in the view of CW. Light yellow and light blue are all instances (50,000) in the view of PGD and CW, while dark yellow and dark blue are top-20\% vulnerable instances (10,000) respectively in the view of PGD and CW. Different distributions of dark yellow and dark blue between subfigure (b) and (c) (or between subfigure (e) and (f)) show that, the vulnerable instances in the view of different attacks are not the same. 
        } 
    \label{fig:moti_2}
    \end{center}
    \vspace{-0.8em}
\end{figure*}

The difference between Eq.~(\ref{Equ-AT-1}) and Eq.~(\ref{Equ-GAIRAT}) is the addition of the geometry-aware weight $\omega(x, y)$. According to GAIRAT \citep{zhang2020geometry}, more vulnerable instances should be assigned larger weights. However, the relative vulnerability between instances may vary in different situations, such as for different adversarial variants. As shown in Figure~\ref{fig:moti_2}, $\mathcal{V}$ represents the selected variable to measure the vulnerability between the classifier and adversarial variants, and the smaller $\mathcal{V}$, the more vulnerable is the instance, which is formally defined in Section~\ref{Sec:LRAT_real}. 
The dark yellow and dark blue are top-20\% vulnerable instances in the view of PGD and CW, respectively. The frequency distribution of dark yellow in Figure~\ref{fig:moti_2}(b) and dark blue in Figure~\ref{fig:moti_2}(c) is clearly different, and the frequency distribution of dark blue in Figure~\ref{fig:moti_2}(e) and dark yellow in Figure~\ref{fig:moti_2}(f) is different. Namely, the most vulnerable 10,000 PGD adversarial variants and the most vulnerable 10,000 CW adversarial variants are not from the same 10,000 instances. As a consequence of this inconsistency, if the attack simulated in training is \emph{mis-specified}, the weights of IRAT are misleading.



\section{Locally Reweighted Adversarial Training}
\label{Sec:LRAT}

To break the limitation of IRAT and train a robust classifier against various attacks, we propose LRAT in this section, for which we perform local reweighting instead of global/no reweighting.

\subsection{Motivation of LRAT}

\textbf{The Reweighting is Beneficial.}
As suggested by GAIRAT \citep{zhang2020geometry}, the global reweighting indeed improves the robustness when tested on the given attack simulated in training. Figures~\ref{fig:moti_1}(b) and \ref{fig:moti_1}(c) also show that, as a global reweighting, GAIRAT improves the robustness against PGD (when PGD is simulated in training). Thus, when training and testing on the same attack, the rationale that we do not need to pay much attention to an already-safe instance under the attack is significant.

\textbf{Local Reweighting can Take Care of Various Attacks Simultaneously.}
As introduced in Section \ref{GAIRAT_limitation}, there is inconsistent vulnerability between instances in different views. Thus, we should perform \emph{local reweighting inside each pair}, while performing \emph{no global reweighting}---the rationale is to fit the instance itself if it is immune to the attack, but not to skip the pair, in order to \emph{passively} defend different attacks in future. In addition, it is inefficient (practically impossible) to simulate all attacks in training, so a gentle lower bound on the instance weights is necessary to defend against potentially adaptive adversarial attacks.

\subsection{Learning Objective of LRAT}
Let $\omega(\tilde{x}, y)$ be the weight assignment function on the loss of adversarial variant $\tilde{x}$. 
The inner optimization for generating $\tilde{x}$ depends on attacks, such as PGD (Eq.~(\ref{Equ-AT-1})). The outer minimization is:
\begin{align}
\mathop{\min}\limits_{f_\theta \in \mathcal{F}}\frac{1}{n}\sum_{i=1}^n\left(\left[\mathcal{C} -\sum_{j=1}^m\omega(\tilde{x}_{ij}, y_{i})\right]_{+}\ell(f_\theta(x_i),y_i)+\sum_{j=1}^m\omega(\tilde{x}_{ij}, y_{i})\ell(f_\theta(\tilde{x}_{ij}),y_{i})\right),
\label{Equ-LRAT}
\end{align}

where $n$ is the number of instances in one mini-batch; $m$ is the number of used attacks; $\mathcal{C}$ is a constant representing the minimum weight sum of each instance; the notation $[a]_+$ stands for $\max\{a,0\}$. We impose two constraints on our objective Eq.~(\ref{Equ-LRAT}): the first constraint ensures that $\omega(\tilde{x}, y) > 0$ and the second constraint ensures that $\mathcal{C} > 0$. The non-negative coefficient $[\mathcal{C}-\sum_{j=1}^m\omega(\tilde{x}_{ij}, y_{i})]_{+}$ assigns some weight to the natural data term, which serves as a gentle lower bound to avoid discarding instances during training. It can also be seen that different weights are assigned to different adversarial variants, respectively. LRAT \emph{pairs} each instance with its adversarial variants and performs \emph{local reweighting inside each pair}. Figure~\ref{fig:sketch} provides an illustrative schematic of the learning objective of LRAT. If $\omega(\tilde{x}, y) = 1$, $\mathcal{C} < 1$, $m = 1$ and $\tilde{x}$ is generated by PGD, LRAT recovers the SAT \citep{Madry18PGD}, which assigns equal weights to the losses of PGD adversarial variant.

\subsection{Realization of LRAT}
\label{Sec:LRAT_real}
The objective in Eq.~(\ref{Equ-LRAT}) implies the optimization process of an adversarially robust network, with one step generating adversarial variants from natural counterparts and then reweighting loss on them, and one step minimizing the reweighted loss w.r.t. the model parameters $\theta$. 

It is still an open question how to calculate the optimal $\omega$ for different variants. \citet{zhang2020geometry} heuristically design some non-increasing functions $\omega$, such as:
\begin{align}
\omega(x,y) = \frac{(1+\tanh(\lambda+5\times(1-2\times\kappa(x,y)/K)))}{2},
\label{Equ-GAIRAT_exa}
\end{align}

where $\kappa/K \in [0,1]$, $K \in \mathbb{N}^{+}$, and $\lambda \in \mathbb{R}$. $\kappa(x,y)$ is the GD defined as the least steps that the PGD method needs to successfully attack the natural instance.
$K$ is the maximally allowed steps. 

\begin{wrapfigure}{r}{0.45\textwidth}
  \begin{center}
    \includegraphics[width=0.45\textwidth]{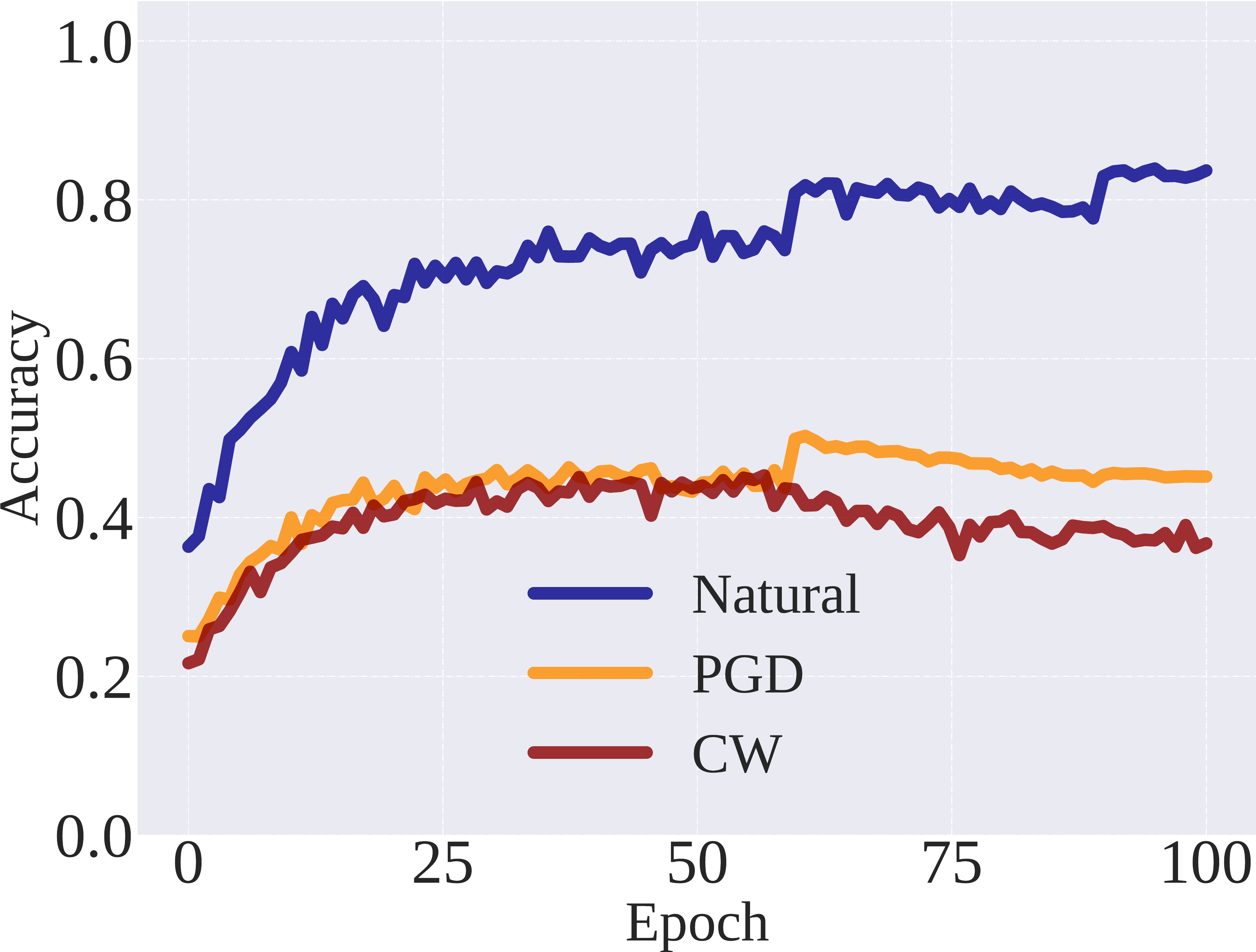}
  \end{center}
  \caption{The limitation of Eq.~(\ref{Equ-GAIRAT_exa}). The figure illustrates the performance on the GAIRAT trained model when the CW is simulated in training.
  } 
  \label{fig:cw}
\end{wrapfigure}
However, the efficacy of this heuristic reweighting function is limited to PGD. When CW is simulated in training, Figure~\ref{fig:cw} shows the same decrease as in Figure~\ref{fig:moti_1} (c) when tested on CW.

Therefore, in this section, we propose a general
\emph{vulnerability-based} and \emph{attack-dependent} reweighting strategy to calculate the weights of the corresponding variants: 
\begin{align}
\label{U:P}
\omega(\tilde{x}, y) \eqdef g(\mathcal{V}_{(\tilde{x}, y)}),
\end{align}
where $\mathcal{V}$ is a predetermined function that measures the vulnerability between the classifier and a certain adversarial variant, and $g$ is a decreasing function of the variable $\mathcal{V}$.

The generations of adversarial variants follow different rules under different attacks. For example, the adversarial variant generated by PGD misleads the classifier with the lowest probability of predicting the correct label \citep{Madry18PGD}. In contrast, the adversarial variant generated by CW misleads the classifier with the highest probability of predicting a wrong label \citep{carlini2017towards}. Motivated by these different generating processes, we define the vulnerability in the view of PGD and CW in the following. 

\begin{algorithm}[!t]
\caption{Locally Reweighted Adversarial Training (LRAT)}
\label{alg:LRATtrain}
\begin{algorithmic}
\STATE \textbf{Input:} network architecture parametrized by $\theta$, training dataset $S$, learning rate $\eta$, number of epochs $T$, batch size $n$, number of batches $N$, number of attacks $m$;
\STATE \textbf{Output:} Adversarial robust network $f_\theta$;
\FOR{$epoch = 1,2,\dots,T$}
\FOR{mini-batch = 1,2,\dots,$N$}
\STATE Sample a mini-batch $\left\{ \left ( x_i,y_i \right) \right\} ^n_{i=1}$ from $S$;
\FOR{$i = 1,2,\dots,n $}
\FOR{$j = 1,2,\dots,m $}
\STATE Obtain adversarial data $\tilde {x}_{ij}$ of ${x}_i$ (e.g., PGD by Algorithm~\ref{alg:pgd});
\STATE Calculate ${w}_{ij}$ according to $\mathcal{V}_{(\tilde{x}_{ij}, y_i)}$ by Eq.~(\ref{U:P});
\ENDFOR
\ENDFOR
\STATE $\theta \gets \theta - \eta  \sum_{i=1}^{n}  \nabla_{\theta} \left[ \left [ \mathcal{C} - \sum_{j=1}^{m} {w}_{ij}\right]_{+} \ell \left ( f_\theta \left ( {x}_i \right),y_i \right) + \sum_{j=1}^{m} {w}_{ij} \ell \left ( f_\theta \left ( \tilde {x}_{ij} \right), y_i \right)\right]/n$;
\ENDFOR
\ENDFOR
\end{algorithmic}
\end{algorithm}

\begin{mydef}[Vulnerability in the view of PGD]
In the view of PGD, the vulnerability  $\mathcal{V}^{\textnormal{PGD}}$ regarding $\tilde{x}$ (generated by PGD) is defined as
\begin{align}
\label{omega:PGD}
\mathcal{V}^{\textnormal{PGD}}_{(\tilde{x}, y)} \eqdef \mathrm{p}_y(\tilde{x}),
\end{align}
where $\mathrm{p}_y$ denotes the predicted probability (softmax on logits) of $\tilde{x}$ belonging to the true class $y$.
\end{mydef}
For the PGD-based adversarial variant, the lower predicted probability it has on true class, the smaller $\mathcal{V}^{\textnormal{PGD}}$ in Eq.~(\ref{omega:PGD}).
\begin{mydef}[Vulnerability in the view of CW]
In the view of CW, the vulnerability $\mathcal{V}^{\textnormal{CW}}$ regarding $\tilde{x}$ (generated by CW) is defined as
\begin{align}
\label{omega:CW}
\mathcal{V}^{\textnormal{CW}}_{(\tilde{x}, y)} \eqdef \mathrm{p}_y(\tilde{x})-\max[\mathrm{p}_i(\tilde{x}):i\neq y],
\end{align}
where $\mathrm{p}_y$ denotes the predicted probability (softmax on logits) of $\tilde{x}$ (CW) belonging to the true class $y$. The $\max[\mathrm{p}_i(\tilde{x}):i\neq y]$ denotes the maximum predicted probability (softmax on logits) of $\tilde{x}$ (CW) belonging to the false class $i$ ($i \neq y$).
\end{mydef}
For the CW-based adversarial variant, the relatively higher predicted probability it has on the false class, the larger $\mathcal{V}^{\textnormal{CW}}$ in Eq.~(\ref{omega:CW}).
In this paper, we consider the decreasing function $g$ with the form
\begin{align}
\label{P:framework}
g(\mathcal{V}) \eqdef \alpha{(1-\mathcal{V})}^{\beta},
\end{align}
where $\alpha > 0$ and $\beta > 0$ are hyper-parameters of each attack. In Eq.~(\ref{P:framework}), adversarial variants with higher $\mathcal{V}$ are given lower weights.
We present our locally reweighted adversarial training in Algorithm~\ref{alg:LRATtrain}. LRAT uses different given attacks (e.g., PGD, Algorithm~\ref{alg:pgd} in Appendix~\ref{App:attack}) to obtain different adversarial variants, and leverages the \emph{attack-dependent} reweighting strategy for obtaining their corresponding weights. For each mini-batch, LRAT reweights the loss of different adversarial variants according to our \emph{vulnerability-based} reweighting strategy, and then updates the model parameters by minimizing the sum of the reweighted loss on each instance.

\begin{table}[!t]
\setlength{\tabcolsep}{5.2mm}
\small
\renewcommand\arraystretch{1.4}
\centering
\caption{Test accuracy (\%) of LRAT and other methods. }
\label{table:LRAT}
\begin{tabular}{c|c|c|c|c}
\toprule[1.5pt]
Methods & Natural & PGD & CW & AA \\
\midrule[0.6pt]
\midrule[0.6pt]
\multicolumn{5}{c}{ResNet-18} \\
\midrule[0.6pt]
\midrule[0.6pt]
AT & ~~~\textbf{82.88}~~~ & ~~~51.16 $\pm$ 0.13~~~ & ~~~49.74 $\pm$ 0.17~~~ & ~~~48.57 $\pm$ 0.11~~~\\
\midrule    
GAIRAT & ~~~80.97~~~ & ~~~\textbf{56.29} $\pm$ 0.19~~~ & ~~~45.77 $\pm$ 0.13~~~ & ~~~32.57 $\pm$ 0.15~~~\\
\midrule    
LRAT & ~~~82.80~~~ & ~~~53.01 $\pm$ 0.13~~~ & ~~~\textbf{50.49} $\pm$ 0.16~~~  & ~~~\textbf{48.60} $\pm$ 0.20~~~\\
\midrule[0.6pt]
\midrule[0.6pt] 
\multicolumn{5}{c}{WRN-32-10} \\
\midrule[0.6pt]
\midrule[0.6pt]
AT & ~~~\textbf{83.42}~~~ & ~~~53.13 $\pm$ 0.18~~~ & ~~~52.26 $\pm$ 0.14~~~ & ~~~\textbf{46.21} $\pm$ 0.14~~~\\
\midrule    
GAIRAT & ~~~82.11~~~ & ~~~\textbf{62.74} $\pm$ 0.08~~~ & ~~~44.63 $\pm$ 0.18~~~ & ~~~44.63 $\pm$ 0.18~~~\\
\midrule    
LRAT & ~~~83.02~~~ & ~~~55.01 $\pm$ 0.19~~~ & ~~~\textbf{53.72} $\pm$ 0.26~~~  & ~~~46.13 $\pm$ 0.15~~~\\
\bottomrule[1.5pt]
\end{tabular}
\vskip1ex%
\vskip -0ex%
\vspace{-1em}
\end{table}\
\begin{table}[!t]
\setlength{\tabcolsep}{2.77mm}
\small
\renewcommand\arraystretch{1.4}
\centering
\caption{Test accuracy (\%) of LRAT-TRADES and other methods. }
\label{table:LRAT-Trades}
\begin{tabular}{c|c|c|c|c|c|c|c|c}
\toprule[1.5pt]
Methods & \multicolumn{2}{|c|}{Natural} &  \multicolumn{2}{|c|}{PGD} & \multicolumn{2}{|c|}{CW} & \multicolumn{2}{|c}{AA} \\
\midrule[0.6pt]
\midrule[0.6pt]
\multicolumn{9}{c}{ResNet-18} \\
\midrule[0.6pt]
\midrule[0.6pt]
Epoch & 60th & 100th & 60th & 100th & 60th & 100th & 60th & 100th \\
\midrule    
TRADES & \makecell{\textbf{78.90}} & \makecell{\textbf{83.15}} & \makecell{53.26 \\ $\pm$ 0.17} & \makecell{52.62 \\ $\pm$ 0.19} & \makecell{51.74 \\ $\pm$ 0.16} & \makecell{50.38 \\ $\pm$ 0.18} & \makecell{48.15 \\ $\pm$ 0.12} & \makecell{47.80 \\ $\pm$ 0.11} \\
\midrule    
GAIR-TRADES  & \makecell{78.55} & \makecell{82.19} & \makecell{\textbf{60.90} \\ $\pm$ 0.18} & \makecell{\textbf{58.17} \\ $\pm$ 0.13} & \makecell{43.39 \\ $\pm$ 0.16} & \makecell{41.27 \\ $\pm$ 0.19} & \makecell{35.29 \\ $\pm$ 0.14} & \makecell{33.24 \\ $\pm$ 0.18} \\
\midrule    
LRAT-TRADES  & \makecell{78.83} & \makecell{82.91} & \makecell{55.21 \\ $\pm$ 0.15} & \makecell{54.77 \\ $\pm$ 0.17} & \makecell{\textbf{52.97} \\ $\pm$ 0.23} & \makecell{\textbf{52.09} \\ $\pm$ 0.14} & \makecell{\textbf{48.24} \\ $\pm$ 0.19} & \makecell{\textbf{47.81} \\ $\pm$ 0.20} \\
\midrule[0.6pt]
\midrule[0.6pt]
\multicolumn{9}{c}{WRN-32-10} \\
\midrule[0.6pt]
\midrule[0.6pt]
Epoch & 60th & 100th & 60th & 100th & 60th & 100th & 60th & 100th \\
\midrule    
TRADES & \makecell{\textbf{82.96}} & \makecell{\textbf{86.21}} & \makecell{55.11 \\ $\pm$ 0.16} & \makecell{54.27 \\ $\pm$ 0.18} & \makecell{54.19 \\ $\pm$ 0.19} & \makecell{53.09 \\ $\pm$ 0.13} & \makecell{52.14 \\ $\pm$ 0.12} & \makecell{51.70 \\ $\pm$ 0.15} \\
\midrule    
GAIR-TRADES  & \makecell{82.20} & \makecell{85.35} & \makecell{\textbf{63.34} \\ $\pm$ 0.17} & \makecell{\textbf{61.27} \\ $\pm$ 0.19} & \makecell{45.31 \\ $\pm$ 0.12} & \makecell{43.32 \\ $\pm$ 0.14} & \makecell{37.82 \\ $\pm$ 0.13} & \makecell{35.88 \\ $\pm$ 0.09} \\
\midrule    
LRAT-TRADES  & \makecell{82.74} & \makecell{85.99} & \makecell{57.68 \\ $\pm$ 0.13} & \makecell{56.89 \\ $\pm$ 0.19} & \makecell{\textbf{55.12} \\ $\pm$ 0.23} & \makecell{\textbf{54.27} \\ $\pm$ 0.17} & \makecell{\textbf{52.17} \\ $\pm$ 0.14} & \makecell{\textbf{51.78} \\ $\pm$ 0.24} \\
\bottomrule[1.5pt]
\end{tabular}
\vskip1ex%
\vskip -0ex%
\end{table}\
\section{Experiments}
\label{Sec:Exp}
In this section, we justify the efficacy of LRAT using networks with various model capacity. In the experiments, we consider $L_{\infty}$-norm bounded perturbation that $||\tilde{x}-x{||}_\infty \leq \epsilon$ in both training and evaluations. All images of the CIFAR-10 are normalized into [0,1]. 

\subsection{Baselines}
We compare LRAT with the no-reweighting strategy (i.e., SAT \citep{Madry18PGD}) and the global-reweighting strategy (i.e., GAIRAT \citep{zhang2020geometry}). 
\citet{rice2020overfitting} show that, unlike in standard training, overfitting in robust adversarial training decays test set performance during training. Thus, as suggested by \citet{rice2020overfitting}, we compare different methods on the performance of the best checkpoint model (the early stopping results at epoch 60). Besides, we also design LRAT for TRADES \citep{ZhangYJXGJ19TRADES}, denoted as (LRAT-TRADES), and the details of the algorithm are in Appendix~\ref{App:LRAT-TRADES}. Accordingly, we also compare LRAT-TRADES with TRADES \citep{ZhangYJXGJ19TRADES} and GAIR-TRADES \citep{zhang2020geometry}. Trades-based methods effectively mitigate the overfitting \citet{rice2020overfitting}, so we compare different methods on both the best checkpoint model and the last checkpoint model (used by \citet{Madry18PGD}), respectively. 

\subsection{Experimental Setup}
We employ the small-capacity network, ResNet (ResNet-18) \citep{he2016deep}, and the large-capacity network, Wide ResNet (WRN-32-10) \citep{zagoruyko2016wide}. Our experimental setup follows previous works \citep{Madry18PGD,wang2019improving,zhang2020geometry}. All networks are trained for 100 epochs using SGD with $0.9$ momentum. The initial learning rate is $0.01$, divided by $10$ at epoch $60$ and $90$, respectively. The weight decay is 0.0035. For generating the PGD adversarial data for updating the network, $L_{\infty}$-norm bounded perturbation $\epsilon_{train} = 8/255$; the maximum PGD step $K = 10$; step size $\alpha = \epsilon_{train} / 10$. For generating the CW adversarial data \citep{carlini2017towards}, we follow the setup in \citep{cai2018curriculum,zhang2020attacks}, where the confidence $\kappa = 50$, and other hyper-parameters are the same as that of PGD above. Robustness to adversarial data is the main evaluation indicator in adversarial training \citep{carlini2019evaluating,chen2020adversarial,cohen2019certified,du2021learning,pang2019improving,yang2021adversarial,zhu2021understanding}. Thus, we evaluate the robust models based on four evaluation metrics, i.e., standard test accuracy on natural data (Natural), robust test accuracy on adversarial data generated by projected gradient descent attack (PGD) \citep{Madry18PGD}, Carlini and Wagner attack (CW) \citep{carlini2017towards} and AutoAttack (AA). In testing, $L_{\infty}$-norm bounded perturbation $\epsilon_{test} = 8/255$, the maximum PGD step $K = 20$, and step size $\alpha = \epsilon / 4$. There is a random start in training and testing, i.e., uniformly random perturbations ($[-\epsilon_{train}, +\epsilon_{train}]$ and $[-\epsilon_{test}, +\epsilon_{test}]$) added to natural instances. Due to the random start, we test our methods and baselines five times with different random seeds.

\subsection{Performance Evaluation}
Tables~\ref{table:LRAT} and \ref{table:LRAT-Trades} report the medians and standard deviations of the results. In our experiments of LRAT, we simulate the PGD and CW in training ($m = 2$ in Eq.~(\ref{Equ-LRAT})). For adversarial variants, we use our \emph{vulnerability-based} reweighting strategy (Eq.~(\ref{U:P})) to obtain their corresponding weights, where $\mathcal{V}^{\textnormal{PGD}}$ follows Eq.~(\ref{omega:PGD}) and $\mathcal{V}^{\textnormal{CW}}$ follows Eq.~(\ref{omega:CW}). We choose the three hyper-parameters ($\mathcal{C}$ in Eq.~(\ref{Equ-LRAT}), $\alpha, \beta$ in Eq.~(\ref{P:framework})) that $\alpha = 2$, $\beta = 0.5$, and $\mathcal{C} = 0.1$, and we analyze it in Appendix~\ref{App:hyper}. 

Compared with SAT, LRAT significantly boosts adversarial robustness under PGD and CW, and the efficacy of LRAT does not diminish under AA. The results show that our local-reweighting strategy is superior to the no-reweighting strategy. Compared with GAIRAT, LRAT has a great improvement under CW and AA. Although GAIRAT improves the performance under PGD, it is not enough to be an effective adversarial defense. The adversarial defense could be described as a barrel Effect. For example, if there is a short slab, the barrel would leak. Similarly, if there is a weakness in defending against some attacks, the classifier would fail to predict. In contrast, LRAT reduces the threat of any potential attacks as an effective defense. Thus, the results also show that our local-reweighting strategy is superior to the global-reweighting strategy. In general, the results affirmatively confirm the efficacy of LRAT. We admit that LRAT has minor improvement under AA (AA does not simulate in training), and the reason behind the limited improvement is the inconsistent vulnerability in different views. Since it is impractical to simulate all attacks in training, and thus we recommend that practitioners simulate multiple attacks and assign some weight to the natural (or PGD adversarial) data term for each instance during training. 
\begin{table}[!t]
\setlength{\tabcolsep}{4.40mm}
\renewcommand\arraystretch{1.4}
\small
\centering
\caption{Objective functions of ablation study.}
\label{table:ablation_form}
\begin{tabular}{c|c|c|c}
\toprule[1.5pt]
Symbol & Objective function & Symbol & Objective function\\
\midrule[0.6pt]
\midrule[0.6pt]
(P) & ~~~$\omega(\tilde{x}^{p}, y)\ell(f_\theta(\tilde{x}^{p}),y)$ ~~~&
([N]) & $\left[\mathcal{C} -\omega(\tilde{x}^{p}, y)-\omega(\tilde{x}^{c}, y)\right]_{+}
\ell(f_\theta({x}),y)$\\
\midrule   
(C) & ~~~$\omega(\tilde{x}^{c}, y)\ell(f_\theta(\tilde{x}^{c}),y)$ ~~~&
([P]) & $\left[\mathcal{C} -\omega(\tilde{x}^{p}, y)-\omega(\tilde{x}^{c}, y)\right]_{+}
\ell(f_\theta(\tilde{x}^{p}),y)$\\
\bottomrule[1.5pt]
\end{tabular}
\end{table}
\begin{table}[!t]
\setlength{\tabcolsep}{4.95mm}
\small
\renewcommand\arraystretch{1.4}
\centering
\caption{Ablation study of LRAT : Test accuracy (\%) using ResNet-18. }
\label{table:ablation_RN-18}
\begin{tabular}{c|c|c|c|c}
\toprule[1.5pt]
Ablation & Natural &  PGD & CW & AA \\
\midrule[0.6pt]
\midrule[0.6pt]
(P) & ~~~80.21~~~ & ~~~52.66 $\pm$ 0.14~~~ & ~~~49.20 $\pm$ 0.22~~~ & ~~~47.72$\pm$ 0.09~~~\\
\midrule    
(C) & ~~~79.55~~~ & ~~~50.57 $\pm$ 0.10~~~ & ~~~\textbf{51.13} $\pm$ 0.16~~~ & ~~~47.81 $\pm$ 0.12~~~\\
\midrule    
(P+C) & ~~~82.40~~~ & ~~~\textbf{53.52} $\pm$ 0.15~~~ & ~~~50.71 $\pm$ 0.08~~~  & ~~~47.80 $\pm$ 0.17~~~\\
\midrule    
([N]+P+C) & ~~~\textbf{82.80}~~~ & ~~~53.01 $\pm$ 0.13~~~ & ~~~50.49 $\pm$ 0.16~~~  & ~~~48.60 $\pm$ 0.20~~~\\
\midrule  
([P]+P+C) & ~~~82.08~~~ & ~~~53.33 $\pm$ 0.18~~~ & ~~~50.60 $\pm$ 0.19~~~  & ~~~\textbf{48.90} $\pm$ 0.12~~~\\
\bottomrule[1.5pt]
\end{tabular}
\vskip1ex%
\vskip -0ex%
\end{table}\
\subsection{Ablation Study}
This subsection validates that each component in LRAT can improve the adversarial robustness. P (or C) represents that only PGD (or CW) with its corresponding reweighting strategy is simulated during training. [N] (or [P]) represents assigning some weights to the natural (or PGD adversarial) data, which serves as a gentle lower bound to avoid discarding instances during training. The objective functions are in Table~\ref{table:ablation_form}, where $\tilde{x}^{p}$ is the adversarial data under PGD, and $\tilde{x}^{c}$ is the adversarial data under CW. The results are reported in Table~\ref{table:ablation_RN-18}. Results show that the robustness of global reweighting (P) and (C) is lower than that of the other three local reweighting when tested on attacks \emph{different from} the given attack simulated during training. Results also show that the robustness of ([N]+P+C) and ([P]+P+C) is higher than that of the other three with no lower bounds when tested on AA, which confirms that a gentle lower bound on instance weights is useful for defending against potentially adaptive adversarial attacks.
\section{Conclusion}

It has been showing great potential to improve adversarial robustness by reweighting adversarial variants during AT. This paper provides a new perspective to this promising direction and aims to train a robust classifier to defend against various attacks. Our proposal, locally reweighted adversarial training (LRAT), pairs each instance with its adversarial variants and performs local reweighting inside each pair. LRAT will not skip any pairs during adversarial training such that it can passively defend against different attacks in future. Experiments show that LRAT works better than both IRAT (i.e., global reweighting) and the standard AT (i.e., no reweighting) when trained with an attack and tested on different attacks. As a general framework, LRAT provides insights on how to design powerful reweighted adversarial training under any potentially adversarial attacks.



\bibliography{example_paper}

\begin{thebibliography}{40}
\providecommand{\natexlab}[1]{#1}
\providecommand{\url}[1]{\texttt{#1}}
\expandafter\ifx\csname urlstyle\endcsname\relax
  \providecommand{\doi}[1]{doi: #1}\else
  \providecommand{\doi}{doi: \begingroup \urlstyle{rm}\Url}\fi

\bibitem[Bai et~al.(2019)Bai, Feng, Wang, Dai, Xia, and Jiang]{bai2019hilbert}
Y.~Bai, Y.~Feng, Y.~Wang, T.~Dai, S.-T. Xia, and Y.~Jiang.
\newblock Hilbert-based generative defense for adversarial examples.
\newblock In \emph{ICCV}, 2019.

\bibitem[Balunovic and Vechev(2019)]{balunovic2019adversarial}
M.~Balunovic and M.~Vechev.
\newblock Adversarial training and provable defenses: Bridging the gap.
\newblock In \emph{ICLR}, 2019.

\bibitem[Cai et~al.(2018)Cai, Du, Liu, and Song]{cai2018curriculum}
Q.-Z. Cai, M.~Du, C.~Liu, and D.~Song.
\newblock Curriculum adversarial training.
\newblock In \emph{IJCAI}, 2018.

\bibitem[Carlini and Wagner(2017{\natexlab{a}})]{carlini2017adversarial}
N.~Carlini and D.~Wagner.
\newblock Adversarial examples are not easily detected: Bypassing ten detection
  methods.
\newblock In \emph{Proceedings of the 10th ACM Workshop on Artificial
  Intelligence and Security}, 2017{\natexlab{a}}.

\bibitem[Carlini and Wagner(2017{\natexlab{b}})]{carlini2017towards}
N.~Carlini and D.~Wagner.
\newblock Towards evaluating the robustness of neural networks.
\newblock In \emph{CVPR}, 2017{\natexlab{b}}.

\bibitem[Carlini et~al.(2019)Carlini, Athalye, Papernot, Brendel, Rauber,
  Tsipras, Goodfellow, Madry, and Kurakin]{carlini2019evaluating}
N.~Carlini, A.~Athalye, N.~Papernot, W.~Brendel, J.~Rauber, D.~Tsipras,
  I.~Goodfellow, A.~Madry, and A.~Kurakin.
\newblock On evaluating adversarial robustness.
\newblock \emph{arXiv preprint arXiv:1902.06705}, 2019.

\bibitem[Chen et~al.(2015)Chen, Seff, Kornhauser, and
  Xiao]{chen2015deepdriving}
C.~Chen, A.~Seff, A.~Kornhauser, and J.~Xiao.
\newblock Deepdriving: Learning affordance for direct perception in autonomous
  driving.
\newblock In \emph{ICCV}, 2015.

\bibitem[Chen et~al.(2020)Chen, Liu, Chang, Cheng, Amini, and
  Wang]{chen2020adversarial}
T.~Chen, S.~Liu, S.~Chang, Y.~Cheng, L.~Amini, and Z.~Wang.
\newblock Adversarial robustness: From self-supervised pre-training to
  fine-tuning.
\newblock In \emph{CVPR}, 2020.

\bibitem[Cohen et~al.(2019)Cohen, Rosenfeld, and Kolter]{cohen2019certified}
J.~Cohen, E.~Rosenfeld, and Z.~Kolter.
\newblock Certified adversarial robustness via randomized smoothing.
\newblock In \emph{ICML}, 2019.

\bibitem[Du et~al.(2021)Du, Zhang, Han, Liu, Rong, Niu, Huang, and
  Sugiyama]{du2021learning}
X.~Du, J.~Zhang, B.~Han, T.~Liu, Y.~Rong, G.~Niu, J.~Huang, and M.~Sugiyama.
\newblock Learning diverse-structured networks for adversarial robustness.
\newblock In \emph{ICML}, 2021.

\bibitem[Finlayson et~al.(2019)Finlayson, Bowers, Ito, Zittrain, Beam, and
  Kohane]{finlayson2019adversarial}
S.~G. Finlayson, J.~D. Bowers, J.~Ito, J.~L. Zittrain, A.~L. Beam, and I.~S.
  Kohane.
\newblock Adversarial attacks on medical machine learning.
\newblock \emph{Science}, 2019.

\bibitem[Gao et~al.(2021)Gao, Liu, Zhang, Han, Liu, Niu, and
  Sugiyama]{gao2020maximum}
R.~Gao, F.~Liu, J.~Zhang, B.~Han, T.~Liu, G.~Niu, and M.~Sugiyama.
\newblock Maximum mean discrepancy is aware of adversarial attacks.
\newblock In \emph{ICML}, 2021.

\bibitem[Goodfellow et~al.(2015)Goodfellow, Shlens, and
  Szegedy]{goodfellow2014explaining}
I.~J. Goodfellow, J.~Shlens, and C.~Szegedy.
\newblock Explaining and harnessing adversarial examples.
\newblock In \emph{ICLR}, 2015.

\bibitem[He et~al.(2016)He, Zhang, Ren, and Sun]{he2016deep}
K.~He, X.~Zhang, S.~Ren, and J.~Sun.
\newblock Deep residual learning for image recognition.
\newblock In \emph{CVPR}, 2016.

\bibitem[He et~al.(2018)He, Li, and Song]{he2018decision}
W.~He, B.~Li, and D.~Song.
\newblock Decision boundary analysis of adversarial examples.
\newblock In \emph{ICLR}, 2018.

\bibitem[Kurakin et~al.(2017)Kurakin, Goodfellow, Bengio,
  et~al.]{kurakin2016adversarial}
A.~Kurakin, I.~Goodfellow, S.~Bengio, et~al.
\newblock Adversarial examples in the physical world.
\newblock In \emph{ICLR}, 2017.

\bibitem[Ma et~al.(2021)Ma, Niu, Gu, Wang, Zhao, Bailey, and
  Lu]{ma2021understanding}
X.~Ma, Y.~Niu, L.~Gu, Y.~Wang, Y.~Zhao, J.~Bailey, and F.~Lu.
\newblock Understanding adversarial attacks on deep learning based medical
  image analysis systems.
\newblock \emph{Pattern Recognition}, 2021.

\bibitem[Madry et~al.(2018)Madry, Makelov, Schmidt, Tsipras, and
  Vladu]{Madry18PGD}
A.~Madry, A.~Makelov, L.~Schmidt, D.~Tsipras, and A.~Vladu.
\newblock Towards deep learning models resistant to adversarial attacks.
\newblock In \emph{ICLR}, 2018.

\bibitem[Miyato et~al.(2017)Miyato, Dai, and Goodfellow]{miyato2016adversarial}
T.~Miyato, A.~M. Dai, and I.~Goodfellow.
\newblock Adversarial training methods for semi-supervised text classification.
\newblock In \emph{ICLR}, 2017.

\bibitem[Nguyen et~al.(2015)Nguyen, Yosinski, and Clune]{nguyen2015deep}
A.~Nguyen, J.~Yosinski, and J.~Clune.
\newblock Deep neural networks are easily fooled: High confidence predictions
  for unrecognizable images.
\newblock In \emph{CVPR}, 2015.

\bibitem[Pang et~al.(2019)Pang, Xu, Du, Chen, and Zhu]{pang2019improving}
T.~Pang, K.~Xu, C.~Du, N.~Chen, and J.~Zhu.
\newblock Improving adversarial robustness via promoting ensemble diversity.
\newblock In \emph{ICML}, 2019.

\bibitem[Papernot et~al.(2016)Papernot, McDaniel, Sinha, and
  Wellman]{papernot2016towards}
N.~Papernot, P.~McDaniel, A.~Sinha, and M.~Wellman.
\newblock Towards the science of security and privacy in machine learning.
\newblock \emph{arXiv:1611.03814}, 2016.

\bibitem[Raghunathan et~al.(2020)Raghunathan, Xie, Yang, Duchi, and
  Liang]{raghunathan2020understanding}
A.~Raghunathan, S.~M. Xie, F.~Yang, J.~Duchi, and P.~Liang.
\newblock Understanding and mitigating the tradeoff between robustness and
  accuracy.
\newblock In \emph{ICML}, 2020.

\bibitem[Rice et~al.(2020)Rice, Wong, and Kolter]{rice2020overfitting}
L.~Rice, E.~Wong, and Z.~Kolter.
\newblock Overfitting in adversarially robust deep learning.
\newblock In \emph{ICML}, 2020.

\bibitem[Shafahi et~al.(2020)Shafahi, Najibi, Xu, Dickerson, Davis, and
  Goldstein]{shafahi2020universal}
A.~Shafahi, M.~Najibi, Z.~Xu, J.~Dickerson, L.~S. Davis, and T.~Goldstein.
\newblock Universal adversarial training.
\newblock In \emph{AAAI}, 2020.

\bibitem[Szegedy et~al.(2014)Szegedy, Zaremba, Sutskever, Bruna, Erhan,
  Goodfellow, and Fergus]{szegedy2013intriguing}
C.~Szegedy, W.~Zaremba, I.~Sutskever, J.~Bruna, D.~Erhan, I.~Goodfellow, and
  R.~Fergus.
\newblock Intriguing properties of neural networks.
\newblock In \emph{ICLR}, 2014.

\bibitem[Tsipras et~al.(2019)Tsipras, Santurkar, Engstrom, Turner, and
  Madry]{tsipras2018robustness}
D.~Tsipras, S.~Santurkar, L.~Engstrom, A.~Turner, and A.~Madry.
\newblock Robustness may be at odds with accuracy.
\newblock In \emph{ICLR}, 2019.

\bibitem[Van~der Maaten and Hinton(2008)]{van2008visualizing}
L.~Van~der Maaten and G.~Hinton.
\newblock Visualizing data using t-sne.
\newblock \emph{Journal of machine learning research}, 2008.

\bibitem[Wang et~al.(2020{\natexlab{a}})Wang, Chen, Gui, Hu, Liu, and
  Wang]{wang2020once}
H.~Wang, T.~Chen, S.~Gui, T.-K. Hu, J.~Liu, and Z.~Wang.
\newblock Once-for-all adversarial training: In-situ tradeoff between
  robustness and accuracy for free.
\newblock In \emph{NeurIPS}, 2020{\natexlab{a}}.

\bibitem[Wang et~al.(2019)Wang, Ma, Bailey, Yi, Zhou, and
  Gu]{wang2019convergence}
Y.~Wang, X.~Ma, J.~Bailey, J.~Yi, B.~Zhou, and Q.~Gu.
\newblock On the convergence and robustness of adversarial training.
\newblock In \emph{ICML}, 2019.

\bibitem[Wang et~al.(2020{\natexlab{b}})Wang, Zou, Yi, Bailey, Ma, and
  Gu]{wang2019improving}
Y.~Wang, D.~Zou, J.~Yi, J.~Bailey, X.~Ma, and Q.~Gu.
\newblock Improving adversarial robustness requires revisiting misclassified
  examples.
\newblock In \emph{ICLR}, 2020{\natexlab{b}}.

\bibitem[Wu et~al.(2020)Wu, Xia, and Wang]{wu2020adversarial}
D.~Wu, S.-T. Xia, and Y.~Wang.
\newblock Adversarial weight perturbation helps robust generalization.
\newblock In \emph{NeurIPS}, 2020.

\bibitem[Yang et~al.(2021)Yang, Guo, Wang, and Xu]{yang2021adversarial}
S.~Yang, T.~Guo, Y.~Wang, and C.~Xu.
\newblock Adversarial robustness through disentangled representations.
\newblock In \emph{AAAI}, 2021.

\bibitem[Yang et~al.(2020)Yang, Rashtchian, Zhang, Salakhutdinov, and
  Chaudhuri]{yang2020closer}
Y.-Y. Yang, C.~Rashtchian, H.~Zhang, R.~Salakhutdinov, and K.~Chaudhuri.
\newblock A closer look at accuracy vs. robustness.
\newblock In \emph{NeurIPS}, 2020.

\bibitem[Zagoruyko and Komodakis(2016)]{zagoruyko2016wide}
S.~Zagoruyko and N.~Komodakis.
\newblock Wide residual networks.
\newblock In \emph{BMVC}, 2016.

\bibitem[Zhang et~al.(2019)Zhang, Yu, Jiao, Xing, Ghaoui, and
  Jordan]{ZhangYJXGJ19TRADES}
H.~Zhang, Y.~Yu, J.~Jiao, E.~P. Xing, L.~E. Ghaoui, and M.~I. Jordan.
\newblock Theoretically principled trade-off between robustness and accuracy.
\newblock In \emph{ICML}, 2019.

\bibitem[Zhang et~al.(2020{\natexlab{a}})Zhang, Xu, Han, Niu, Cui, Sugiyama,
  and Kankanhalli]{zhang2020attacks}
J.~Zhang, X.~Xu, B.~Han, G.~Niu, L.~Cui, M.~Sugiyama, and M.~Kankanhalli.
\newblock Attacks which do not kill training make adversarial learning
  stronger.
\newblock In \emph{ICML}, 2020{\natexlab{a}}.

\bibitem[Zhang et~al.(2021)Zhang, Zhu, Niu, Han, Sugiyama, and
  Kankanhalli]{zhang2020geometry}
J.~Zhang, J.~Zhu, G.~Niu, B.~Han, M.~Sugiyama, and M.~Kankanhalli.
\newblock Geometry-aware instance-reweighted adversarial training.
\newblock In \emph{ICLR}, 2021.

\bibitem[Zhang et~al.(2020{\natexlab{b}})Zhang, Li, Liu, and
  Tian]{zhang2020dual}
Y.~Zhang, Y.~Li, T.~Liu, and X.~Tian.
\newblock Dual-path distillation: A unified framework to improve black-box
  attacks.
\newblock In \emph{ICML}, 2020{\natexlab{b}}.

\bibitem[Zhu et~al.(2021)Zhu, Zhang, Han, Liu, Niu, Yang, Kankanhalli, and
  Sugiyama]{zhu2021understanding}
J.~Zhu, J.~Zhang, B.~Han, T.~Liu, G.~Niu, H.~Yang, M.~Kankanhalli, and
  M.~Sugiyama.
\newblock Understanding the interaction of adversarial training with noisy
  labels.
\newblock \emph{arXiv preprint arXiv:2102.03482}, 2021.

\end{thebibliography}
\bibliographystyle{abbrvnat}


\clearpage
\appendix
\label{Appendix}
\section{Adversarial Attack}
\label{App:attack}
Algorithm~\ref{alg:pgd} and Algorithm~\ref{alg:cw} are the adversarial data generation of PGD and CW, respectively. The loss function in PGD is:
\begin{align}
\label{eq:ce}
\ell_{CE} \eqdef -\log{\mathrm{p}_y(\tilde{x})},
\end{align}
where $\mathrm{p}_y$ denotes the predicted probability (softmax on logits) of $\tilde{x}$ belonging to the true class $y$.

The loss function in CW is:
\begin{align}
\label{eq:cw}
\ell_{CW} \eqdef -\mathrm{Z}_y(\tilde{x})+\max[\mathrm{Z}_i(\tilde{x}):i\neq y]-\kappa,
\end{align}
where $\mathrm{Z}_y$ denotes the predicted probability (before softmax) of $\tilde{x}$ belonging to the true class $y$; $\kappa = 50$ (following the setup in \citep{cai2018curriculum,zhang2020attacks}) is a hyper-parameter to denote the confidence.

\section{TRadeoff-inspired Adversarial DEfense via Surrogate-loss Minimization}
\label{App:LRAT-TRADES}

The objective function of TRadeoff-inspired Adversarial DEfense via Surrogate-loss minimization (TRADES) \citep{ZhangYJXGJ19TRADES} is
\begin{align}
\mathop{\min}\limits_{f_\theta \in \mathcal{F}}\frac{1}{n}\sum_{i=1}^n \left( \ell(f_\theta(x_i),y_i)+\frac{1}{\lambda}\ell_{KL}(f_\theta(\tilde{x}_i),f_\theta(x_i))\right),
\end{align}
where 
\begin{align}
\tilde{x}_i =\mathop{\argmax}\limits_{\tilde{x}_i \in \epsball[x]}\ell_{KL}(f_\theta(\tilde{x}_i),f_\theta(x_i)).
\end{align}

$\lambda > 0$ is a regularization parameter. Others remain the same with standard adversarial training. The approximation method for searching adversarial data in TRADES is as follows:
\begin{align}
x^{(t+1)} = \Pi_{\mathcal{B}[x^{(0)}]}(x^{(t)}+\alpha sign(\nabla_{x^{t}}\ell_{KL}(f_{\theta}(\tilde{x}^{(t)}),f_{\theta}(x^{(t)})))), t \in \mathbb{N}.
\end{align}
Instead of the loss function $\ell:\mathbb{R}^{\mathcal{C}}\times \mathcal{Y} \to \mathbb{R}$ in Eq.~(\ref{Equ-AT-3}), TRADES use $\ell_{KL}$ that the $KL$ divergence between the prediction of natural data and their adversarial variants, i.e.:
\begin{align}
\mathcal{R}(x,\delta; \theta) = \mathcal{L}_{KL}[\mathrm{p}(x;\theta)\parallel \mathrm{p}(\tilde{x};\theta)] = \sum_{k}\mathrm{p}_k(x;\theta)\log\frac{\mathrm{p}_k(x;\theta)}{\mathrm{p}_k(\tilde{x}; \theta)}.
\end{align}
TRADES generates the adversatial data with a maximum predicted probability distribution difference from the natural data, instead of generating the most adversarial data in AT. We also design LRAT for TRADES that LRAT-TRADES in Algorithm~\ref{alg:LRAT_TRADES}.

\section{Experimental Details}
\label{App:hyper}
\textbf{Weighting Normalization.} In our experiments, we impose another constraint on our objective Eq.~(\ref{Equ-LRAT}): 
\begin{align}
\label{eq:cons}
\frac{1}{n}\sum_{i=1}^n \left([\mathcal{C}-\sum_{j=1}^m\omega(\tilde{x}_{ij},y_{i})]_{+}+\sum_{j=1}^m\omega(\tilde{x}_{ij}, y_{i})\right) = 1,
\end{align}
to implement a fair comparison with baselines.

\begin{algorithm}[!t]
\small
\footnotesize
\caption{Adversarial Data Generation in Projected Gradient Descent Attack (PGD)}
\label{alg:pgd}
\begin{algorithmic}
\STATE \textbf{Input:} natural data $x \in \mathcal{X}$, label $y \in \mathcal{Y}$, model $f$, loss funciton $\ell_{CE}$, maximum PGD step $K$, 
\STATE perturbation bound $\epsilon$, step size $\alpha$;
\STATE \textbf{Output:} adversarial data $\tilde{x}$;
\STATE$\tilde{x} \gets x$;
\WHILE{$K > 0$} 
\STATE $\tilde{x} \gets \Pi_{\mathcal{B}_{\epsilon}[x]}(\tilde{x}+\alpha sign(\nabla_{\tilde{x}}\ell_{CE}(f(\tilde{x}),y))$;
\STATE $K \gets K - 1$;
\ENDWHILE
\end{algorithmic}
\end{algorithm}

\begin{algorithm}[!t]
\small
\footnotesize
\caption{Adversarial Data Generation in Carlini and Wagner Attack (CW)}
\label{alg:cw}
\begin{algorithmic}
\STATE \textbf{Input:} natural data $x \in \mathcal{X}$, label $y \in \mathcal{Y}$, model $f$, loss funciton $\ell_{CW}$, maximum PGD step $K$, 
\STATE perturbation bound $\epsilon$, step size $\alpha$;
\STATE \textbf{Output:} adversarial data $\tilde{x}$;
\STATE$\tilde{x} \gets x$;
\WHILE{$K > 0$} 
\STATE $\tilde{x} \gets \Pi_{\mathcal{B}_{\epsilon}[x]}(\tilde{x}+\alpha sign(\nabla_{\tilde{x}}\ell_{CW}(f(\tilde{x}),y))$;
\STATE $K \gets K - 1$;
\ENDWHILE
\end{algorithmic}
\end{algorithm}

\begin{algorithm}[!t]
\small
\caption{Locally Reweighted Adversarial Training for TRADES (LRAT-TRADES)}
\label{alg:LRAT_TRADES}
\begin{algorithmic}
\STATE \textbf{Input:} network architecture parametrized by $\theta$, training dataset $S$, learning rate $\eta$, number of epochs $T$, batch size $n$, number of batches $N$, number of attacks $m$;
\STATE \textbf{Output:} Adversarial robust network $f_\theta$;
\FOR{$epoch = 1,2,\dots,T$}
\FOR{mini-batch = 1,2,\dots,$N$}
\STATE Sample a mini-batch $\left\{ \left ( x_i,y_i \right) \right\} ^n_{i=1}$ from $S$;
\FOR{$i = 1,2,\dots,n $}
\FOR{$j = 1,2,\dots,m $}
\STATE Obtain adversarial data $\tilde {x}_{ij}$ of ${x}_i$ (e.g., PGD by Algorithm~\ref{alg:pgd});
\STATE Calculate ${w}_{ij}$ according to $\mathcal{V}_{(\tilde{x}_{ij}, y_i)}$ by Eq.~(\ref{U:P});
\ENDFOR
\ENDFOR
\STATE $\theta \gets \theta - \eta  \sum_{i=1}^{n}  \nabla_{\theta} \left[ \left [ \mathcal{C} - \sum_{j=1}^{m} {w}_{ij}\right]_{+} \ell_{CE} \left ( f_\theta \left ( {x}_i \right),y_i \right) + \sum_{j=1}^{m} {w}_{ij} \ell_{KL} \left ( f_\theta \left ( \tilde {x}_{ij} \right), f_\theta \left ( {x}_{ij} \right) \right)\right]/n$;
\ENDFOR
\ENDFOR
\end{algorithmic}
\end{algorithm}

\textbf{Selection of Hyper-parameters.} 
It is an open question how to define the decreasing function $g$ in Eq.~(\ref{U:P}). (or given the definition of $g$ in Eq.~(\ref{P:framework}), how to select the hyper-parameters under different situations.) In our experiments, given an alternative value set $\{0.25,0.5,1.5,2,4\}$, we choose $\alpha, \beta$ in Eq.~(\ref{P:framework}) that $\alpha = 2$, $\beta = 0.5$, and given an alternative value set $\{0.1,0.2,0.4,0.6\}$, we choose $\mathcal{C}$ in Eq.~(\ref{Equ-LRAT}) that $\mathcal{C} = 0.1$. Our experiments show that the performance has a minor variation between different hyper-parameters, and we choose the optimal hyper-parameters depending on whose robustness attacked by PGD is the best.

\textbf{Natural Data Term.} 
To avoid discarding instances during training, we impose the non-negative coefficient $[\mathcal{C}-\sum_{j=1}^m\omega(\tilde{x}_{ij}, y_{i})]_{+}$ in Eq.~(\ref{Equ-LRAT}) to assign some weight to the natural data term. The definition of the natural data term can vary as required by different tasks. When practitioners only focus on the robustness under attacks simulated in training, this term can be eliminated. When practitioners focus on the robustness under attacks simulated in training and the accuracy on natural data, this term can be defined as the loss of the natural instance.  When practitioners focus on the robustness under attacks both simulated and unseen in training, this term can be defined as the loss of the adversarial variant. In our experiments, natural data term in LRAT (the ([N]+P+C) in our ablation study) is the loss sum of the natural instance and the PGD adversarial variant, which aims to achieve robustness and accuracy.
\section{Experimental Resources}
\label{App:resource}
We implement all methods on Python $3.7$ (Pytorch $1.7.1$) with a NVIDIA GeForce RTX 3090 GPU with AMD Ryzen Threadripper 3960X 24 Core Processor. The CIFAR-10 dataset and the SVHN dataset can be downloaded via Pytorch. See the codes submitted. Given the $50,000$ images from the CIFAR-10 training set and $73,257$ digits from the SVHN training set, we conduct the adversarial training on ResNet-18 and Wide ResNet-32 for classification. DNNS are trained using SGD with $0.9$ momentum, the initial learning rate of $0.01$ and the batch size of $128$ for $100$ epochs. 

\section{Additional Experiments on the SVHN}
\label{App:Experiments}
We also justify the efficacy of LRAT on the SVHN. In the experiments, we employ ResNet-18 and consider $L_{\infty}$-norm bounded perturbation that $||\tilde{x}-x{||}_\infty \leq \epsilon$ in both training and evaluations. All images of the SVHN are normalized into [0,1]. Tables~\ref{table:LRAT_SVHN} reports the medians and standard deviations of the results. Compared with SAT, LRAT significantly boosts adversarial robustness under PGD and CW, and the efficacy of LRAT does not diminish under AA. Compared with GAIRAT, LRAT has a great improvement under CW and AA. 
\begin{table}[!t]
\setlength{\tabcolsep}{5.2mm}
\small
\renewcommand\arraystretch{1.4}
\centering
\caption{Test accuracy (\%) of LRAT and other methods on the SVHN. }
\label{table:LRAT_SVHN}
\begin{tabular}{c|c|c|c|c}
\toprule[1.5pt]
Methods & Natural & PGD & CW & AA \\
\midrule[0.6pt]
\midrule[0.6pt]
AT & ~~~\textbf{92.38}~~~ & ~~~55.97 $\pm$ 0.18~~~ & ~~~52.90 $\pm$ 0.21~~~ & ~~~\textbf{47.84} $\pm$ 0.17~~~\\
\midrule    
GAIRAT & ~~~90.31~~~ & ~~~\textbf{62.96} $\pm$ 0.10~~~ & ~~~47.17 $\pm$ 0.17~~~ & ~~~38.74 $\pm$ 0.19~~~\\
\midrule    
LRAT & ~~~92.30~~~ & ~~~59.76 $\pm$ 0.14~~~ & ~~~\textbf{55.47} $\pm$ 0.18~~~  & ~~~47.65 $\pm$ 0.27~~~\\
\bottomrule[1.5pt]
\end{tabular}
\vskip1ex%
\vskip -0ex%
\vspace{-1em}
\end{table}\
\section{Discussions on the defense against unseen attacks}
\label{App:Discussions}
As a general framework, LRAT provides insights on how to design powerful reweighting adversarial training under different adversarial attacks. Duo to the inconsistent vulnerability in different views, reweighting adversarial training has the risk of weakening the ability to defend against unseen attacks. Thus, we recommend that practitioners simulate diverse attacks during training. Note that it does not mean that practitioners should use different attacks indiscriminately---for instance, during standard adversarial training, mixing some weak adversarial data into PGD adversarial data will weaken the robustness on the contrary. The recommended diversity is diverse information focused on during adversarial data generation, such as the difference between misleading the classifier with the lowest probability of predicting the correct label (PGD) and misleading the classifier with the highest probability of predicting a wrong label (CW).

\end{document}